\documentclass[12pt]{article}
\usepackage[margin=1.25in]{geometry}
\usepackage{amsmath}
\usepackage{times}
\usepackage{graphicx}
\usepackage{color}
\usepackage{multirow}
\usepackage{apacite}
\usepackage[authoryear]{natbib}
\usepackage{rotating}
\usepackage{bbm}
\usepackage{latexsym}

\usepackage{amsmath,amsfonts,bm}

\def\eqref#1{equation~\ref{#1}}
\def\1{\bm{1}}

\DeclareMathAlphabet{\mathsfit}{\encodingdefault}{\sfdefault}{m}{sl}
\SetMathAlphabet{\mathsfit}{bold}{\encodingdefault}{\sfdefault}{bx}{n}

\usepackage{lipsum}
\usepackage{subcaption}
\usepackage{bm}
\usepackage{hyperref}
\usepackage{algorithm}
\usepackage{algorithmicx}
\usepackage{algpseudocode}
\usepackage{amsmath}
\usepackage{amsthm}
\usepackage{amssymb}
\usepackage{amsfonts}
\usepackage{wrapfig}
\usepackage{microtype}      % microtypography
\usepackage{empheq}
\usepackage{multicol}
\usepackage{mathrsfs}
\usepackage{booktabs}

\newcommand{\diag}{\operatorname{diag}}

\newcommand*\widefbox[1]{\fbox{\hspace{2em}#1\hspace{2em}}}

\begin{document}
% \hspace{13.9cm}1

% \ \vspace{20mm}\\
\begin{center}
    {\LARGE Predictive Coding, Variational Autoencoders, \\ and Biological Connections}
\end{center}

\ \\
{\bf \large Joseph Marino}\footnote{Now at DeepMind, London, UK.}\\
\small \texttt{josephmarino@deepmind.com} \\
\normalsize {Computation \& Neural Systems, \\ California Institute of Technology, \\ Pasadena, CA 91125, USA}\\
\\
%

%\ \\[-2mm]

\thispagestyle{empty}
\markboth{}{NC instructions}
\ \vspace{-0mm}\\
%
%Abstract
\begin{center} {\bf Abstract} \end{center}
This paper reviews predictive coding, from theoretical neuroscience, and variational autoencoders, from machine learning, identifying the common origin and mathematical framework underlying both areas. As each area is prominent within its respective field, more firmly connecting these areas could prove useful in the dialogue between neuroscience and machine learning. After reviewing each area, we discuss two possible correspondences implied by this perspective: cortical pyramidal dendrites as analogous to (non-linear) deep networks and lateral inhibition as analogous to normalizing flows. These connections may provide new directions for further investigations in each field.
%%%%%%%%%%%

\section{Introduction}
\label{sec: pc intro}

\subsection{Cybernetics}

Machine learning and theoretical neuroscience once overlapped under the field of cybernetics \citep{wiener1948cybernetics, ashby1956introduction}. Within this field, perception and control, both in biological and non-biological systems, were formulated in terms of negative feedback and feedforward processes. Negative feedback attempts to minimize error signals by \textit{feed}ing the errors \textit{back} into the system, whereas feedforward processing attempts to preemptively reduce error through prediction. Cybernetics formalized these techniques using probabilistic models, which estimate the likelihood of random outcomes, and variational calculus, a technique for estimating functions, particularly probability distributions \citep{wiener1948cybernetics}. This resulted in the first computational models of neuron function and learning \citep{mcculloch1943logical, rosenblatt1958perceptron, widrow1960adaptive}, a formal definition of information \citep{wiener1942interpolation, shannon1948mathematical} (with connections to neural systems \citep{barlow1961coding}), and algorithms for negative feedback perception and control \citep{mackay1956epistemological, kalman1960new}. Yet, with advances in these directions (see \citet{prieto2016neural} and references therein), the cohesion of cybernetics diminished, with the new ideas taking root in theoretical neuroscience, machine learning, control theory, etc.

\subsection{Neuroscience \& Machine Learning: Convergence \& Divergence}

A renewed dialogue between neuroscience and machine learning formed in the 1980s--1990s. Neuroscientists, bolstered by new physiological and functional analyses, began making traction in studying neural systems in probabilistic and information-theoretic terms \citep{laughlin1981simple, srinivasan1982predictive, barlow1989unsupervised, bialek1991reading}. In machine learning, improvements in probabilistic modeling \citep{pearl1986fusion} and artificial neural networks \citep{rumelhart1986learning} combined with ideas from statistical mechanics \citep{hopfield1982neural, ackley1985learning} to yield new classes of models and training techniques. This convergence of ideas, primarily centered around perception, resulted in new theories of neural processing and improvements in their mathematical underpinnings.

In particular, the notion of \textit{predictive coding} emerged within neuroscience \citep{srinivasan1982predictive, rao1999predictive}. In its most general form, predictive coding postulates that neural circuits are engaged in estimating probabilistic models of other neural activity and sensory inputs, with feedback and feedforward processes playing a central role. These models were initially formulated in early sensory areas, e.g.,~retina \citep{srinivasan1982predictive} and thalamus \citep{dong1995temporal}, using feedforward processes to predict future neural activity. Similar notions were extended to higher-level sensory processing in neocortex by David Mumford \citep{mumford1991computational, mumford1992computational}. Top-down neural projections (from higher-level to lower-level sensory areas) were hypothesized to convey sensory predictions, whereas bottom-up neural projections were hypothesized to convey prediction errors. Through negative feedback, these errors then updated state estimates. These ideas were formalized by \cite{rao1999predictive}, formulating a simplified artificial neural network model of images, reminiscent of a Kalman filter \citep{kalman1960new}.

\begin{figure}[t!]
    \centering
    \includegraphics[width=0.9\textwidth]{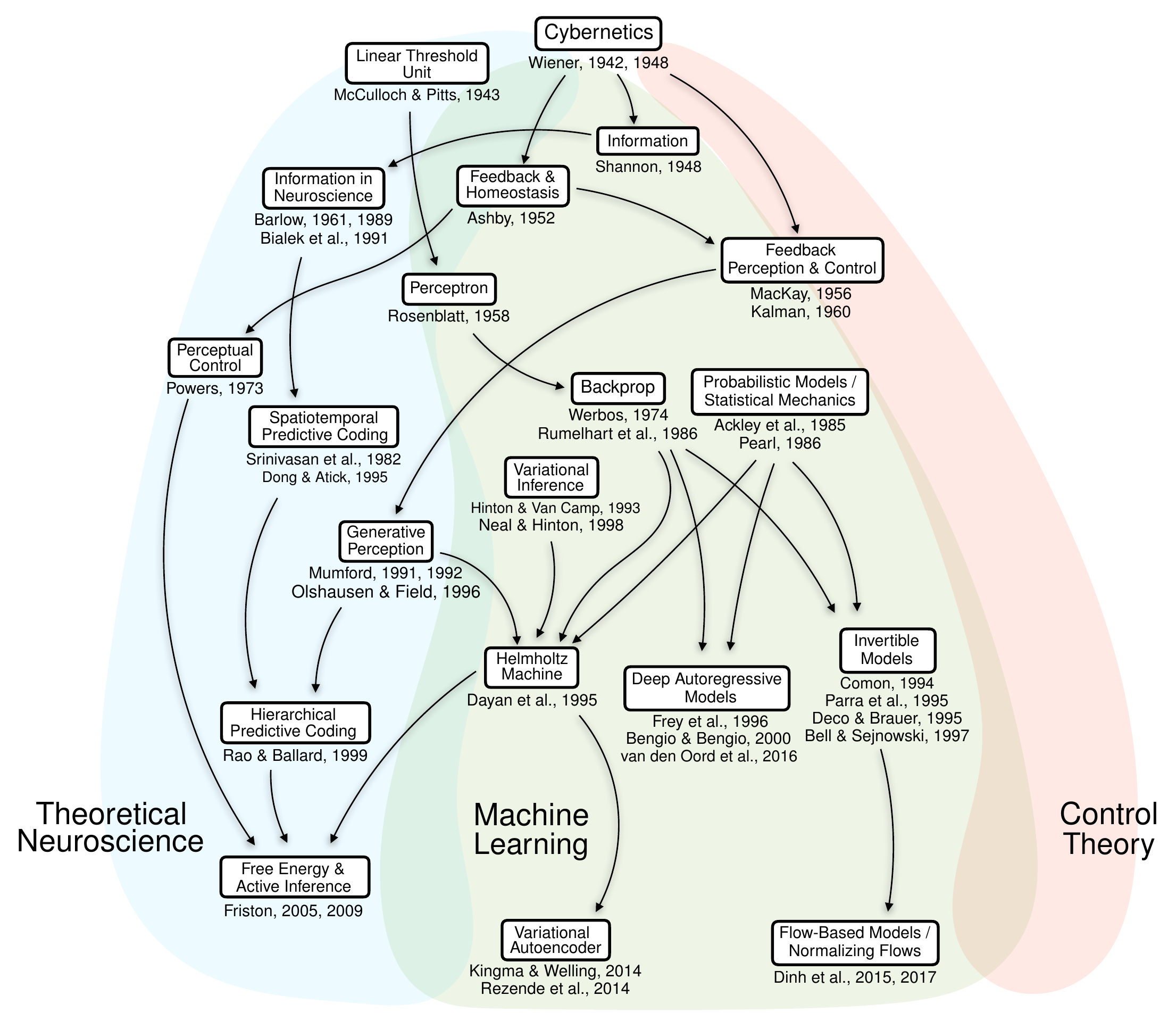}
    \caption{\textbf{Concept Overview}. Cybernetics influenced the areas that became theoretical neuroscience and machine learning, resulting in shared mathematical concepts. This paper explores the connections between predictive coding, from theoretical neuroscience, and variational autoencoders, from machine learning.}
    \label{fig: concept map}
\end{figure}

Feedback and feedforward processes also featured prominently in machine learning. Indeed, the primary training algorithm for artificial neural networks, backpropagation \citep{rumelhart1986learning}, literally \textit{feeds} (propagates) the output prediction errors \textit{back} through the network, i.e.,~negative feedback. During this period, the technique of variational inference was rediscovered within machine learning \citep{hinton1993keeping, neal1998view}, recasting probabilistic inference using variational calculus. This technique proved essential in formulating the Helmholtz machine \citep{dayan1995helmholtz, dayan1996varieties}, a hierarchical unsupervised probabilistic model parameterized by artificial neural networks. Similar advances were made in autoregressive probabilistic models \citep{frey1996does, bengio2000modeling}, using artificial neural networks to form sequential feedforward predictions, as well as new classes of invertible probabilistic models \citep{comon1994independent, parra1995redundancy, deco1995higher, bell1997independent}.
% Yet, with the ``AI winter'' of the late 1990s, progress in these areas slowed.

These new ideas regarding variational inference and probabilistic models, particularly the Helmholtz machine \citep{dayan1995helmholtz}, influenced predictive coding. Specifically, Karl Friston utilized variational inference to formulate hierarchical dynamical models of neocortex \citep{friston2005theory, friston2008hierarchical}. In line with \citet{mumford1992computational}, these models contain multiple levels, with each level attempting to predict its future activity (feedforward) as well as lower-level activity, closer to the input data. Prediction errors across levels facilitate updating higher-level estimates (negative feedback). Such models have incorporated many biological aspects, including local learning rules \citep{friston2005theory} and attention \citep{spratling2008reconciling, feldman2010attention, kanai2015cerebral}, and have been compared with neural circuits \citep{bastos2012canonical, keller2018predictive, walsh2020evaluating}. While predictive coding and other Bayesian brain theories are increasingly popular \citep{doya2007bayesian, friston2009free, clark2013whatever}, validating these models is hampered by the difficulty of distinguishing between specific design choices and general theoretical claims \citep{gershman2019does}. Further, a large gap remains between the simplified implementations of these models and the complexity of neural systems.

Progress in machine learning picked up in the early 2010s, with advances in parallel computing as well as standardized datasets \citep{deng2009imagenet}. In this era of \textit{deep learning} \citep{lecun2015deep, schmidhuber2015deep}, i.e.,~artificial neural networks with multiple layers, a flourishing of ideas emerged around probabilistic modeling. Building off of previous work, more expressive classes of deep hierarchical \citep{gregor2013deep, mnih2014neural, kingma2014stochastic, rezende2014stochastic}, autoregressive \citep{uria2014deep, van2016pixel}, and invertible \citep{dinh2015nice, dinh2017density} probabilistic models were developed. Of particular importance is a model class known as \textit{variational autoencoders} (VAEs) \citep{kingma2014stochastic, rezende2014stochastic}, a relative of the Helmholtz machine, which closely resembles hierarchical predictive coding. Unfortunately, despite this similarity, the machine learning community remains largely oblivious to the progress in predictive coding and vice versa.

\subsection{Connecting Predictive Coding and VAEs}

This paper aims to bridge the divide between predictive coding and VAEs. While the present work provides unique contributions, it is inspired by previous works at this intersection. In particular, \cite{van2016auto} outlines hierarchical probabilistic models in predictive coding and machine learning. Likewise, \cite{lotter2016deep, lotter2018neural} implement predictive coding techniques in deep probabilistic models, comparing these models with neural phenomena.

After reviewing background mathematical concepts (Sec.~\ref{sec: background}), we discuss the basic formulations of predictive coding (Sec.~\ref{sec: predictive coding}) and variational autoencoders (Sec.~\ref{sec: vaes}), identifying commonalities in their model formulations and inference techniques (Sec.~\ref{sec: pc connections}). Based on these connections, in Sec.~\ref{sec: pc correspondences}, we discuss two possible correspondences between machine learning and neuroscience seemingly suggested by this perspective:
\begin{itemize}
    \item \textbf{dendrites of pyramidal neurons \& deep artificial networks}, affirming a more nuanced perspective over the analogy of biological and artificial neurons, and
    \item \textbf{lateral inhibition \& normalizing flows}, providing a more general framework for normalization.
\end{itemize}
Like the works of \cite{van2016auto} and \cite{lotter2016deep, lotter2018neural}, we hope that these connections will inspire future research in exploring this promising direction.

\section{Background}
\label{sec: background}

\subsection{Maximum Log-Likelihood}

Consider a random variable, $\mathbf{x} \in \mathbb{R}^M$, with a corresponding distribution, $p_{\small \textrm{data}} (\mathbf{x})$, defining the probability of observing each possible value. This distribution is the result of an underlying data generating process, e.g., the emission and scattering of photons. While we do not have direct access to $p_{\small \textrm{data}}$, we can sample observations, $\mathbf{x} \sim p_{\small \textrm{data}} (\mathbf{x})$, yielding an empirical distribution, $\widehat{p}_{\small \textrm{data}} (\mathbf{x})$. Often, we wish to model $p_{\small \textrm{data}}$, e.g., for prediction or compression. We refer to this model as $p_\theta (\mathbf{x})$, with parameters $\theta$. Estimating the model parameters involves maximizing the log-likelihood of data samples under the model's distribution:
\begin{equation}
    \theta^* \leftarrow \textrm{arg} \max_\theta \mathbb{E}_{\mathbf{x} \sim p_\textrm{data} (\mathbf{x})} \left[ \log p_\theta (\mathbf{x}) \right]. \label{eq: max likelihood 3}
\end{equation}
This is the \textit{maximum log-likelihood} objective, which is found throughout machine learning and probabilistic modeling \citep{murphy2012machine}. In practice, we do not have access to $p_{\small \textrm{data}} (\mathbf{x})$ and instead approximate the objective using data samples, i.e., using $\widehat{p}_{\small \textrm{data}} (\mathbf{x})$.

\subsection{Probabilistic Models}

\subsubsection{Dependency Structure}
\label{sec: background dep struct}

A probabilistic model includes the dependency structure (Sec.~\ref{sec: background dep struct}) and the parameterization of these dependencies (Sec.~\ref{sec: background parameterizing the model}). The dependency structure is the set of conditional dependencies between variables. One common form is given by \textbf{autoregressive models} \citep{frey1996does, bengio2000modeling}, which use the chain rule of probability:
\begin{equation}
    p_\theta (\mathbf{x}) = \prod_{j=1}^M p_\theta (x_j | x_{<j}).
    \label{eq: background ar non-temporal}
\end{equation}
By inducing an ordering over the $M$ dimensions of $\mathbf{x}$, we can factor the joint distribution, $p_\theta (\mathbf{x})$, into a product of $M$ conditional distributions, each conditioned on the previous dimensions, $x_{<j}$. A natural use-case arises in modeling sequential data, where time provides an ordering over a sequence of $T$ variables, $\mathbf{x}_{1:T}$:
\begin{equation}
    p_\theta (\mathbf{x}_{1:T}) = \prod_{t=1}^T p_\theta (\mathbf{x}_t | \mathbf{x}_{<t}).
    \label{eq: background ar temporal}
\end{equation}
% Although such models are conventionally formulated in forward temporal order, this is a modeling assumption. Likewise, while Eq.~\ref{eq: background ar temporal} considers \textit{all} previous variables, there may be cases where it is safe to assume conditional independence outside of some window.

\begin{figure}[t!]
    \centering
    \includegraphics[width=0.9\textwidth]{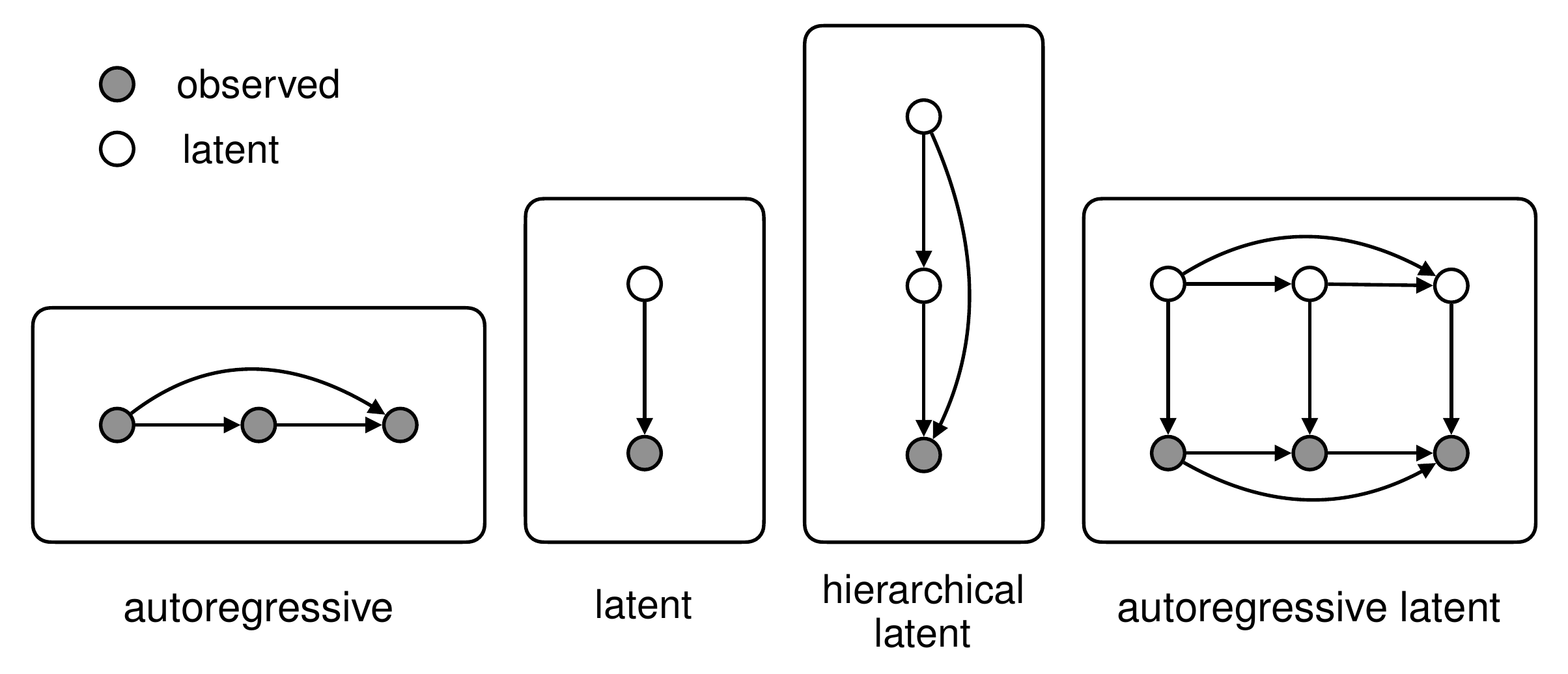}
    \caption{\textbf{Dependency Structures}. Each diagram shows a directed graphical model. Nodes represent random variables, and arrows represent dependencies. The main forms of dependency structure are autoregressive (Eq.~\ref{eq: background ar non-temporal}) and latent variable models (Eq.~\ref{eq: background basic lvm factorization}). These structures can be combined in various ways (Eqs.~\ref{eq: background hierarchical lvm} \& \ref{eq: background sequential lvm}).}
    \label{fig: dependency structure}
\end{figure}

Autoregressive models are ``fully-visible'' models \citep{frey1996does}, as dependencies are only modeled between observed variables. However, we can also introduce \textit{latent variables}, $\mathbf{z}$. Formally, a \textbf{latent variable model} is defined by the joint distribution
\begin{equation}
    p_\theta (\mathbf{x}, \mathbf{z} ) = p_\theta (\mathbf{x} | \mathbf{z}) p_\theta (\mathbf{z}),
    \label{eq: background basic lvm factorization}
\end{equation}
where $p_\theta (\mathbf{x} | \mathbf{z})$ is the \textit{conditional likelihood} and $p_\theta (\mathbf{z})$ is the \textit{prior}. Introducing latent variables is one of, if not, \textit{the} primary technique for increasing the flexibility of a model, as evaluating the likelihood now requires marginalizing over the latent variables:
% If $\mathbf{z}$ is a continuous variable, this involves integration, $p_\theta (\mathbf{x}) = \int p_\theta (\mathbf{x}, \mathbf{z}) d \mathbf{z}$, and if $\mathbf{z}$ is discrete, this involves summation, $p_\theta (\mathbf{x}) = \sum_{\mathbf{z}} p_\theta (\mathbf{x}, \mathbf{z})$. In either case, we have
\begin{equation}
    p_\theta (\mathbf{x}) = \mathbb{E}_{\mathbf{z} \sim p_\theta (\mathbf{z})} \left[ p_\theta (\mathbf{x} | \mathbf{z}) \right].
    \label{eq: background marginalization}
\end{equation}
Thus, $p_\theta (\mathbf{x})$ is a \textit{mixture} distribution, with each component, $p_\theta (\mathbf{x} | \mathbf{z})$, weighted according to $p_\theta (\mathbf{z})$. Even when $p_\theta (\mathbf{x} | \mathbf{z})$ takes a simple distribution form, such as Gaussian, $p_\theta (\mathbf{x})$ can take on flexible forms. In this way, $\mathbf{z}$ can capture complex dependencies in $\mathbf{x}$.

However, increased flexibility comes with increased computational cost. In general, marginalizing over $\mathbf{z}$ is not tractable. This requires us to 1) adopt approximations, discussed in Sec.~\ref{sec: background variational inference}, or 2) restrict the model to ensure tractable evaluation of $p_\theta (\mathbf{x})$ (Eq.~\ref{eq: background marginalization}). The latter approach is taken by \textbf{flow-based latent variable models} \citep{tabak2013family, rippel2013high, dinh2015nice}, defining the dependency between $\mathbf{x}$ and $\mathbf{z}$ via an invertible transform, $\mathbf{x} = f_\theta (\mathbf{z})$ and $\mathbf{z} = f_\theta^{-1} (\mathbf{x})$. With a prior or \textit{base distribution}, $p_\theta (\mathbf{z})$, we can express $p_\theta (\mathbf{x})$ using the \textit{change of variables formula}:
\begin{equation}
    p_\theta (\mathbf{x}) = p_\theta (\mathbf{z}) \left| \det \left(\frac{\partial \mathbf{x}}{\partial \mathbf{z}} \right) \right|^{-1},
    \label{eq: background change of variables}
\end{equation}
where $\frac{\partial \mathbf{x}}{\partial \mathbf{z}}$ is the Jacobian of the transform and $\det(\cdot)$ denotes matrix determinant. The term $\left| \det \left(\frac{\partial \mathbf{x}}{\partial \mathbf{z}} \right) \right|^{-1}$ is the local scaling of space when moving from $\mathbf{z}$ to $\mathbf{x}$, conserving probability mass. Flow-based models, also referred to as \textit{normalizing flows} \citep{rezende2015variational}, are the basis of independent components analysis (ICA) \citep{comon1994independent, bell1997independent, hyvarinen2000independent} and non-linear generalizations \citep{chen2001gaussianization, laparra2011iterative}. These models provide a general-purpose mechanism for adding and removing dependencies between variables, i.e., normalization.\footnote{\normalsize Formally, we refer to normalization as one or more steps of a process transforming the data density into a standard Gaussian (i.e., Normal), which is equivalent to ICA \citep{hyvarinen2000independent}. This is a form of redundancy reduction, removing statistical dependencies between data dimensions.} Yet, while flow-based models avoid marginalization, their requirement of invertibility can be overly restrictive \citep{cornish2020relaxing}.

We have presented autoregression and latent variables separately, however, these techniques can be combined. For instance, \textit{hierarchical latent variable models} \citep{dayan1995helmholtz} incorporate autoregressive dependencies between latent variables. Considering $L$ levels of latent variables, $\mathbf{z}^{1:L} = \left[\mathbf{z}^1, \dots, \mathbf{z}^L  \right]$, we can express the joint distribution as
\begin{equation}
    p_\theta (\mathbf{x}, \mathbf{z}^{1:L}) = p_\theta (\mathbf{x} | \mathbf{z}^{1:L}) \prod_{\ell = 1}^L p_\theta (\mathbf{z}^\ell | \mathbf{z}^{\ell+1 : L}).
    \label{eq: background hierarchical lvm}
\end{equation}
% Hierarchical latent variable models create increasingly abstract \textit{empirical priors} \citep{efron1973stein}.
We can also incorporate latent variables within sequential (autoregressive) models, giving rise to \textit{sequential latent variable models}. Considering a single level of latent variables in a corresponding sequence, $\mathbf{z}_{1:T}$, we have the following joint distribution:
\begin{equation}
    p_\theta (\mathbf{x}_{1:T}, \mathbf{z}_{1:T}) = \prod_{t=1}^T p_\theta (\mathbf{x}_t | \mathbf{x}_{<t}, \mathbf{z}_{\leq t}) p_\theta (\mathbf{z}_t | \mathbf{x}_{<t}, \mathbf{z}_{<t}).
    \label{eq: background sequential lvm}
\end{equation}
This encompasses special cases, such as hidden Markov models or linear state-space models \citep{murphy2012machine}. There are a variety of other ways to combine autoregression and latent variables \citep{gulrajani2017pixelvae, razavi2019generating}. In some cases, autoregressive and flow-based latent variable models are even equivalent \citep{kingma2016improved}.

\subsubsection{Parameterizing the Model}
\label{sec: background parameterizing the model}

The distributions defining probabilistic dependencies are functions. In this section, we discuss forms that these functions may take. The canonical example is the Gaussian (or Normal) distribution, $\mathcal{N} (x; \mu, \sigma^2)$, which is defined by a mean, $\mu$, and variance, $\sigma^2$. This can be extended to the multivariate setting, where $\mathbf{x} \in \mathbb{R}^M$ is modeled with a mean vector, $\bm{\mu}$, and covariance matrix, $\bm{\Sigma}$, with the probability density written as
\begin{equation}
    \mathcal{N} (\mathbf{x}; \bm{\mu}, \bm{\Sigma}) = \frac{1}{(2 \pi)^{M/2} \det (\bm{\Sigma})^{1/2}} \exp \left( \frac{-1}{2} (\mathbf{x} - \bm{\mu})^\intercal \bm{\Sigma}^{-1} (\mathbf{x} - \bm{\mu}) \right).
    \label{eq: gaussian def}
\end{equation}
To simplify calculations, we may consider diagonal covariance matrices, $\bm{\Sigma} = \diag (\bm{\sigma}^2)$. In particular, the special case where $\bm{\Sigma} = \mathbf{I}_M$, the $M \times M$ identity matrix, the log-density becomes the \textit{mean squared error}:
\begin{equation}
    \log \mathcal{N} (\mathbf{x}; \bm{\mu}, \mathbf{I}) = -\frac{1}{2} || \mathbf{x} - \bm{\mu} ||^2_2 + \textrm{const.}
\end{equation}

Conditional dependencies are mediated by the distribution parameters, which are functions of the conditioning variables. For example, we can express an autoregressive Gaussian distribution (Eq.~\ref{eq: background ar non-temporal}) through $p_\theta (x_j | x_{<j}) = \mathcal{N} (x_j; \mu_\theta (x_{<j}), \sigma^2_\theta  (x_{<j}))$, where $\mu_\theta$ and $\sigma^2_\theta$ are functions taking $x_{<j}$ as input. A similar form applies to autoregressive models on sequences of vector inputs (Eq.~\ref{eq: background ar temporal}), with $p_\theta (\mathbf{x}_t | \mathbf{x}_{<t}) = \mathcal{N} (\mathbf{x}_t; \bm{\mu}_\theta (\mathbf{x}_{<t}), \bm{\Sigma}_\theta (\mathbf{x}_{<t}))$. Likewise, in a latent variable model (Eq.~\ref{eq: background basic lvm factorization}), we can express a Gaussian conditional likelihood as $p_\theta ( \mathbf{x} | \mathbf{z}) = \mathcal{N} (\mathbf{x}; \bm{\mu}_\theta (\mathbf{z}), \bm{\Sigma}_\theta (\mathbf{z}))$. In the above examples, we have used a subscript $\theta$ for all functions, however, these may be separate functions in practice.

The functions supplying each of the distribution parameters can range in complexity, from constant to highly non-linear. Classical modeling techniques often employ linear functions. For instance, in a latent variable model, we could parameterize the mean as
\begin{equation}
    \bm{\mu}_\theta (\mathbf{z}) = \mathbf{W} \mathbf{z} + \mathbf{b},
    \label{eq: background linear mean lvm}
\end{equation}
where $\mathbf{W}$ is a matrix of weights and $\mathbf{b}$ is a bias vector. Models of this form underlie factor analysis, probabilistic principal components analysis \citep{tipping1999probabilistic}, independent components analysis \citep{bell1997independent, hyvarinen2000independent}, and sparse coding \citep{olshausen1996emergence}. While linear models are computationally efficient, they are often too limited to accurately model complex data distributions, e.g., those found in natural images or audio.

Deep learning \citep{goodfellow2016deep} provides probabilistic models with expressive non-linear functions, improving their capacity. In these models, the distribution parameters are parameterized with deep networks, which are then trained by backpropagating \citep{rumelhart1986learning} the gradient of the log-likelihood, $\nabla_\theta \mathbb{E}_{\mathbf{x} \sim \widehat{p}_{\small \textrm{data}}} \left[ \log p_\theta (\mathbf{x}) \right]$, through the network. Deep probabilistic models have enabled recent advances in speech \citep{graves2013generating, van2016wavenet}, natural language \citep{sutskever2014sequence, radford2019language}, images \citep{razavi2019generating}, video \citep{kumar2020videoflow}, reinforcement learning \citep{chua2018deep, ha2018recurrent} and other areas.

\begin{figure}[t!]
    \centering
    \includegraphics[width=0.73\textwidth]{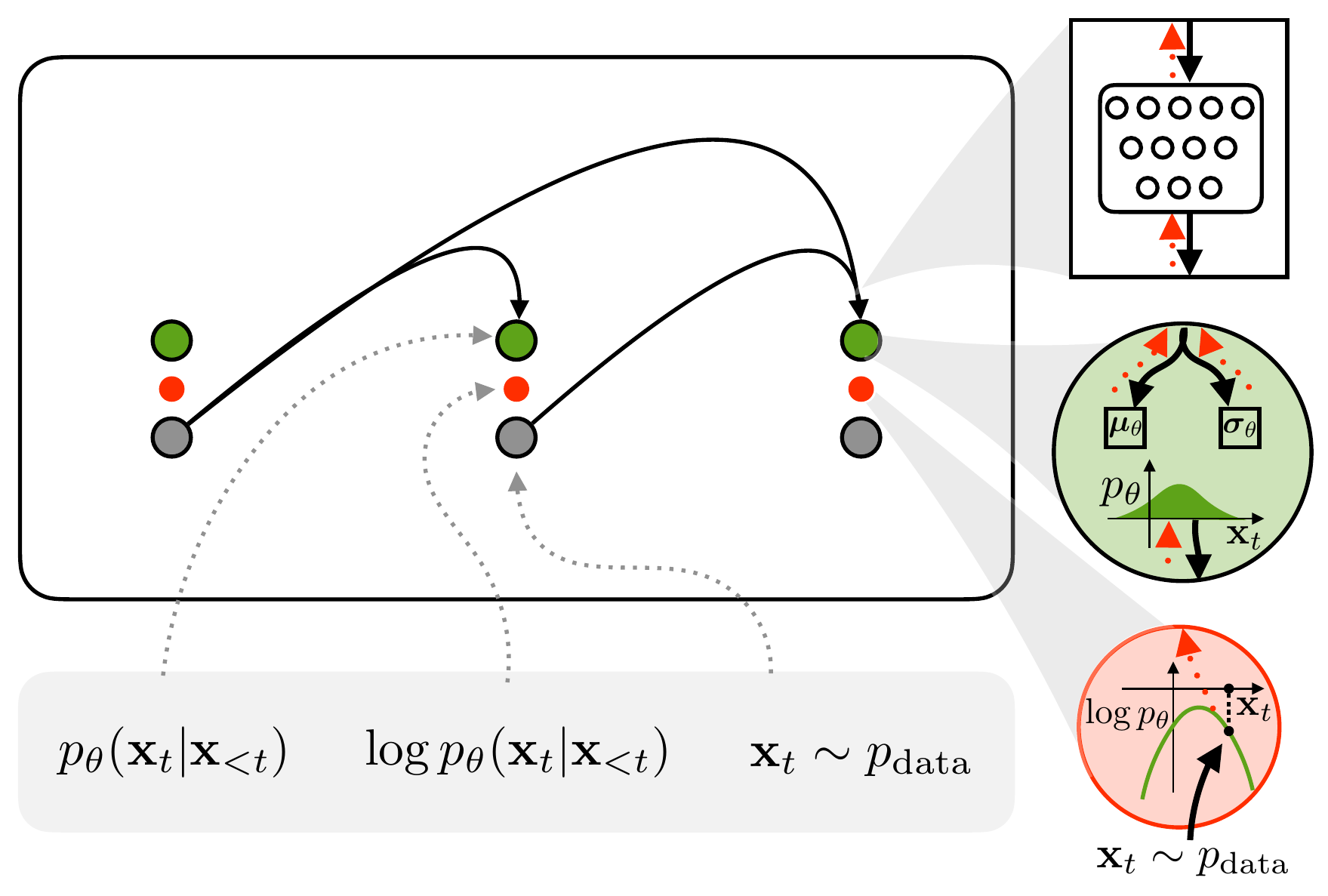}
    \caption{\textbf{Autoregressive Computation Graph}. The graph contains the (Gaussian) conditional likelihoods (green), data (gray), and terms in the objective (red dots). Gradients (red dotted lines) backpropagate through the networks parameterizing the distributions.}
    \label{fig: background model param}
\end{figure}

In Figure~\ref{fig: background model param}, we visualize a computation graph for a deep autoregressive model, breaking the variables into their distributions and terms in the objective. Green circles denote the (Gaussian) conditional likelihoods at each step, which are parameterized by deep networks. The log-likelihood, $\log p_\theta (\mathbf{x}_t | \mathbf{x}_{<t})$, evaluated at the data observation, $\mathbf{x}_t \sim p_\textrm{data} (\mathbf{x}_t | \mathbf{x}_{<t})$ (gray), provides the objective (red dot). The gradient of this objective w.r.t.~the network parameters is calculated through backpropagation (red dotted line).

Autoregressive models have proven useful in many domains. However, there are reasons to prefer latent variable models in some contexts. First, autoregressive sampling is inherently sequential, becoming costly in high-dimensional domains. Second, latent variables provide a representation for downstream tasks, compression, and overall data analysis. Finally, latent variables increase flexibility, which is useful for modeling complex distributions with relatively simple, e.g., Gaussian, conditional distributions. While flow-based latent variable models offer one option, their invertibility requirement limits the types of functions that can be used. For these reasons, we require methods for handling the latent marginalization in Eq.~\ref{eq: background marginalization}. Variational inference is one such method.

\subsection{Variational Inference}
\label{sec: background variational inference}

Training latent variable models through maximum likelihood requires evaluating $\log p_\theta (\mathbf{x})$. However, evaluating $p_\theta (\mathbf{x}) = \int p_\theta (\mathbf{x}, \mathbf{z}) d \mathbf{z}$ is generally intractable. Thus, we require some technique for approximating $\log p_\theta (\mathbf{x})$. Variational inference \citep{hinton1993keeping, jordan1998introduction} approaches this problem by introducing an \textit{approximate posterior} distribution, $q (\mathbf{z} | \mathbf{x})$, which provides a lower bound, $\mathcal{L} (\mathbf{x}; q, \theta) \leq \log p_\theta (\mathbf{x})$. This lower bound is referred to as the evidence (or variational) lower bound (ELBO) as well as the (negative) free energy. By tightening and maximizing the ELBO w.r.t.~the model parameters, $\theta$, we can approximate maximum likelihood training.

Variational inference converts probabilistic inference into an optimization problem. Given a family of distributions, $\mathcal{Q}$, e.g., Gaussian, variational inference attempts to find the distribution, $q \in \mathcal{Q}$, that minimizes $D_\textrm{KL} (q (\mathbf{z} | \mathbf{x}) || p_\theta (\mathbf{z} | \mathbf{x}))$,
% \begin{equation}
%     q (\mathbf{z} | \mathbf{x}) \leftarrow \textrm{arg} \min_q D_\textrm{KL} (q (\mathbf{z} | \mathbf{x}) || p_\theta (\mathbf{z} | \mathbf{x})),
%     \label{eq: vi kl minimization}
% \end{equation}
where $p_\theta (\mathbf{z} | \mathbf{x})$ is the \textit{posterior} distribution, $p_\theta (\mathbf{z} | \mathbf{x}) = \frac{p_\theta (\mathbf{x}, \mathbf{z})}{p_\theta (\mathbf{x})}$. Because $p_\theta (\mathbf{z} | \mathbf{x})$ includes the intractable $p_\theta (\mathbf{x})$, we cannot minimize this KL divergence directly. Instead, we can rewrite this as
\begin{equation}
    D_\textrm{KL} (q (\mathbf{z} | \mathbf{x}) || p_\theta (\mathbf{z} | \mathbf{x})) = \log p_\theta (\mathbf{x}) - \mathcal{L} (\mathbf{x}; q, \theta),
    \label{eq: background elbo deriv rearrange}
\end{equation}
% \begin{align}
%     D_\textrm{KL} (q (\mathbf{z} | \mathbf{x}) || p_\theta (\mathbf{z} | \mathbf{x})) & = \mathbb{E}_{\mathbf{z} \sim q (\mathbf{z} | \mathbf{x})} \left[ \log q (\mathbf{z} | \mathbf{x}) - \log p_\theta (\mathbf{z} | \mathbf{x}) \right]  \nonumber \\
%     & = \mathbb{E}_{\mathbf{z} \sim q (\mathbf{z} | \mathbf{x})} \left[ \log q (\mathbf{z} | \mathbf{x}) - \log \left( \frac{p_\theta (\mathbf{x} , \mathbf{z})}{p_\theta (\mathbf{x})} \right) \right]  \nonumber \\
%     & = \underbrace{\mathbb{E}_{\mathbf{z} \sim q (\mathbf{z} | \mathbf{x})} \left[ \log q (\mathbf{z} | \mathbf{x}) - \log p_\theta (\mathbf{x} , \mathbf{z}) \right]}_{- \mathcal{L} (\mathbf{x}; q, \theta)} + \log p_\theta (\mathbf{x}). \label{eq: background elbo deriv rearrange}
%     % & = - \mathcal{L} (\mathbf{x}; q, \theta) + \log p_\theta (\mathbf{x}). \label{eq: background elbo deriv elbo def}
% \end{align}
where, in Eq.~\ref{eq: background elbo deriv rearrange} (see Appendix~\ref{app: elbo deriv}), we have defined $\mathcal{L} (\mathbf{x}; q, \theta)$, as
\begin{empheq}[box=\widefbox]{align}
    \mathcal{L} (\mathbf{x}; q, \theta) & \equiv \mathbb{E}_{\mathbf{z} \sim q (\mathbf{z} | \mathbf{x})} \left[ \log p_\theta (\mathbf{x} , \mathbf{z}) - \log q (\mathbf{z} | \mathbf{x}) \right] \label{eq: elbo def 1} \\
    & = \mathbb{E}_{\mathbf{z} \sim q (\mathbf{z} | \mathbf{x})} \left[ \log p_\theta (\mathbf{x} | \mathbf{z}) \right] - D_\textrm{KL} (q (\mathbf{z} | \mathbf{x}) || p_\theta (\mathbf{z})). \label{eq: elbo def 2}
\end{empheq}
Rearranging terms in Eq.~\ref{eq: background elbo deriv rearrange}, we have
\begin{equation}
    \log p_\theta (\mathbf{x}) = \mathcal{L} (\mathbf{x}; q, \theta) + D_\textrm{KL} (q (\mathbf{z} | \mathbf{x}) || p_\theta (\mathbf{z} | \mathbf{x})).
    \label{eq: background elbo def 3}
\end{equation}
Because KL divergence is non-negative, $\mathcal{L} (\mathbf{x}; q, \theta) \leq \log p_\theta (\mathbf{x})$, with equality when $q (\mathbf{z} | \mathbf{x}) = p_\theta (\mathbf{z} | \mathbf{x})$. As the LHS of Eq.~\ref{eq: background elbo def 3} does not depend on $q (\mathbf{z} | \mathbf{x})$, maximizing $\mathcal{L} (\mathbf{x}; q, \theta)$ w.r.t.~$q$ implicitly minimizes $D_\textrm{KL} (q (\mathbf{z} | \mathbf{x}) || p_\theta (\mathbf{z} | \mathbf{x}))$ w.r.t.~$q$. Thus, maximizing $\mathcal{L} (\mathbf{x}; q, \theta)$ w.r.t.~$q$ tightens the lower bound on $\log p_\theta (\mathbf{x})$. With this tightened lower bound, we can then maximize $\mathcal{L} (\mathbf{x}; q, \theta)$ w.r.t.~$\theta$. This alternating optimization process is the variational expectation maximization (EM) algorithm \citep{dempster1977maximum, neal1998view}, consisting of approximate inference (E-step) and learning (M-step).

\begin{figure}[t!]
    \centering
    \begin{subfigure}[t]{0.44\textwidth}
        \centering
        \includegraphics[width=0.90\textwidth]{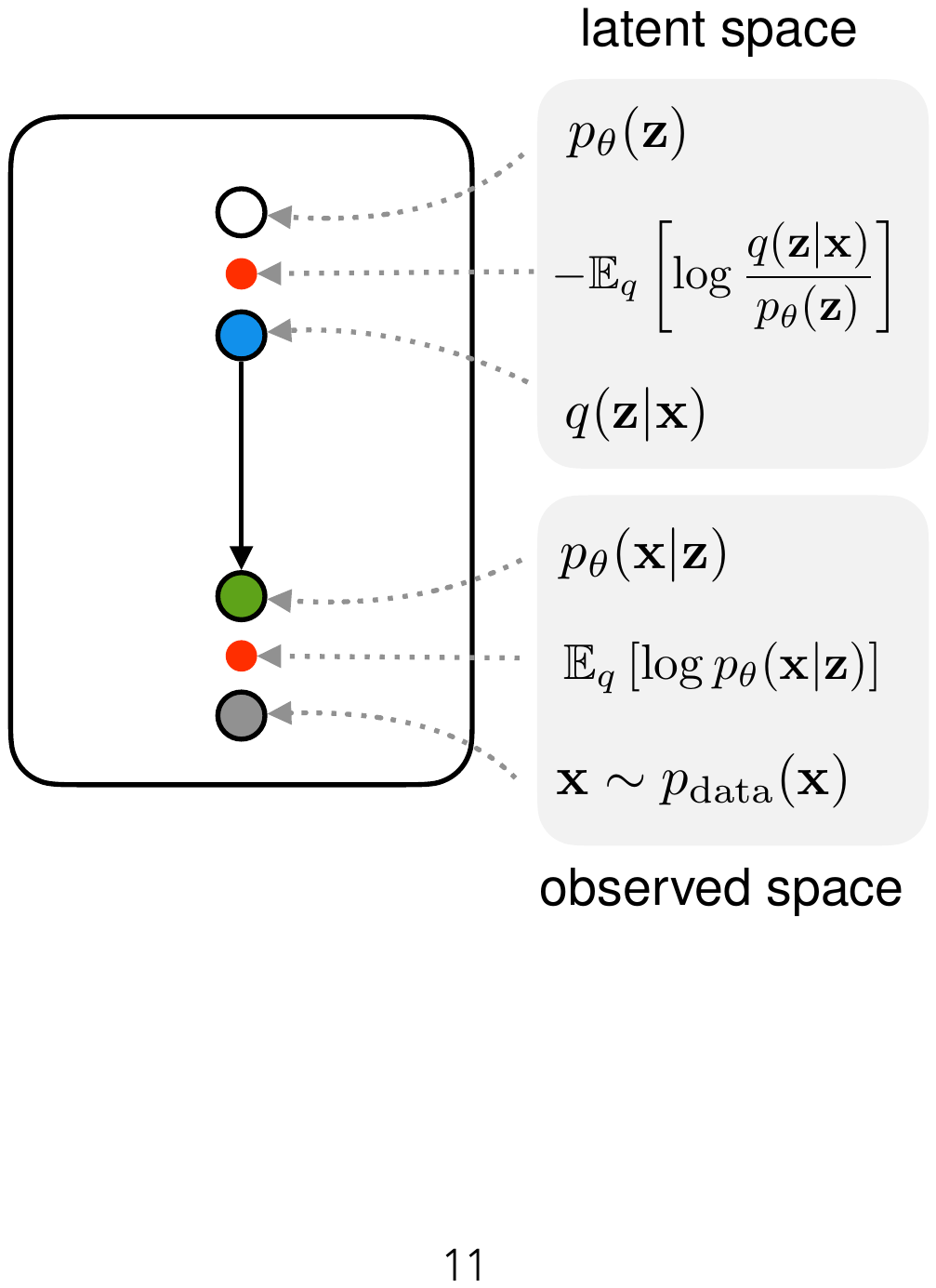}
        \caption{}
        \label{fig: background elbo comp graph}
    \end{subfigure}%
    ~ 
    \begin{subfigure}[t]{0.14\textwidth}
        \centering
        \includegraphics[width=\textwidth]{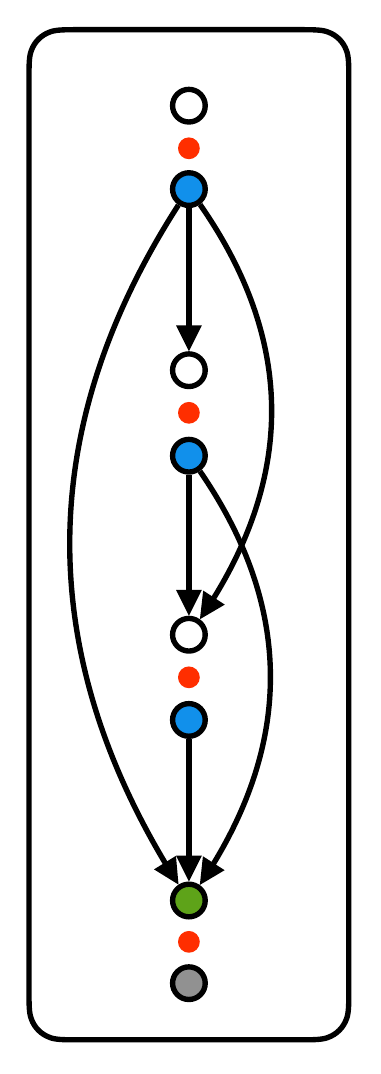}
        \caption{}
        \label{fig: background elbo comp graph hierarchical}
    \end{subfigure}%
    ~ 
    \begin{subfigure}[t]{0.41\textwidth}
        \centering
        \includegraphics[width=\textwidth]{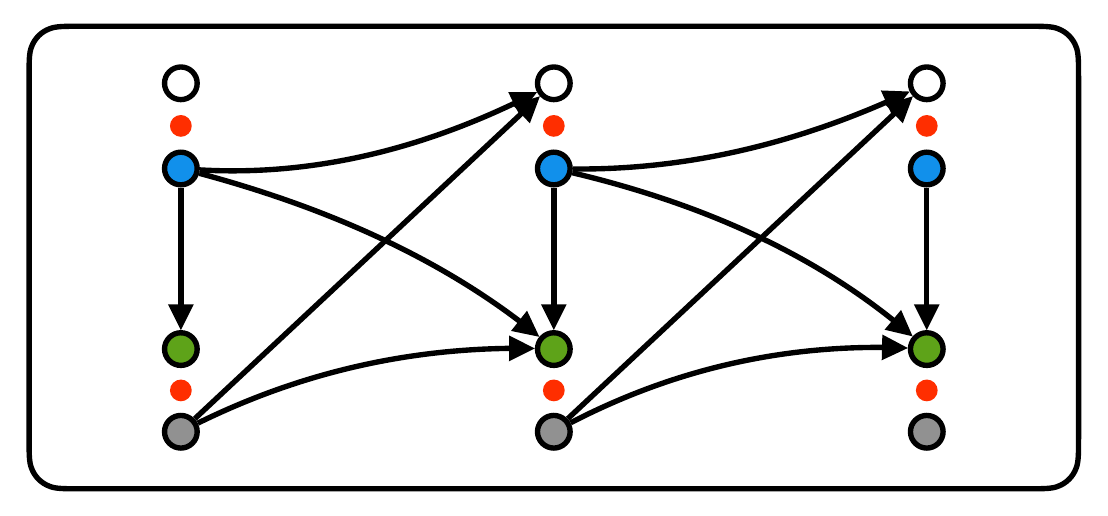}
        \caption{}
        \label{fig: background elbo comp graph sequential}
    \end{subfigure}
    \caption{\textbf{ELBO Computation Graphs}. \textbf{(a)} Basic computation graph for variational inference. Outlined circles denote distributions, smaller red circles denote terms in the ELBO, and arrows denote conditional dependencies. This notation can be used to express \textbf{(b)} hierarchical and  \textbf{(c)} sequential models with various dependencies.}
    \label{fig: background elbo comp graphs}
\end{figure}

% \begin{figure}[t!]
%     \centering
%     \begin{subfigure}[t]{0.37\textwidth}
%         \centering
%         \includegraphics[width=\textwidth]{figures/elbo_comp_graph.pdf}
%         \caption{Computation Graph}
%         \label{fig: background elbo comp graph}
%     \end{subfigure}%
%     ~ 
%     \begin{subfigure}[t]{0.27\textwidth}
%         \centering
%         \includegraphics[width=\textwidth]{figures/elbo_comp_graph_hierarchical.pdf}
%         \caption{Hierarchical}
%         \label{fig: background elbo comp graph hierarchical}
%     \end{subfigure}%
%     ~ 
%     \begin{subfigure}[t]{0.34\textwidth}
%         \centering
%         \includegraphics[width=\textwidth]{figures/elbo_comp_graph_sequential.pdf}
%         \caption{Sequential}
%         \label{fig: background elbo comp graph sequential}
%     \end{subfigure}
%     \caption{\textbf{ELBO Computation Graphs}. \textbf{(a)} Basic computation graph for variational inference. Outlined circles denote distributions. Smaller red circles denote terms in the ELBO. Arrows denote conditional dependencies. This notation can be used to express \textbf{(b)} hierarchical and  \textbf{(c)} sequential models with various dependencies.}
%     \label{fig: background elbo comp graphs}
% \end{figure}

We can also represent the ELBO in latent variable models as a computation graph (Figure~\ref{fig: background elbo comp graphs}). Each variable contains a red circle, denoting a term in the ELBO. As compared with the log-likelihood objective, we now have an additional objective term corresponding to the KL divergence for the latent variable. We visualize the variational objective for more complex hierarchical and sequential models in Figures~\ref{fig: background elbo comp graph hierarchical}~\&~\ref{fig: background elbo comp graph sequential}.

\section{Predictive Coding}
\label{sec: predictive coding}

Predictive coding can be divided into two settings, spatiotemporal and hierarchical, roughly corresponding to the two main forms of probabilistic dependencies. In this section, we review these settings, discussing existing hypothesized correspondences with neural anatomy. We then outline the empirical support for predictive coding, highlighting the need for large-scale, testable models. 

\subsection{Spatiotemporal Predictive Coding}
\label{sec: pc spatiotemporal}

\begin{figure}[t!]
    \centering
    \begin{subfigure}[t]{0.58\textwidth}
    \includegraphics[width=0.97\textwidth]{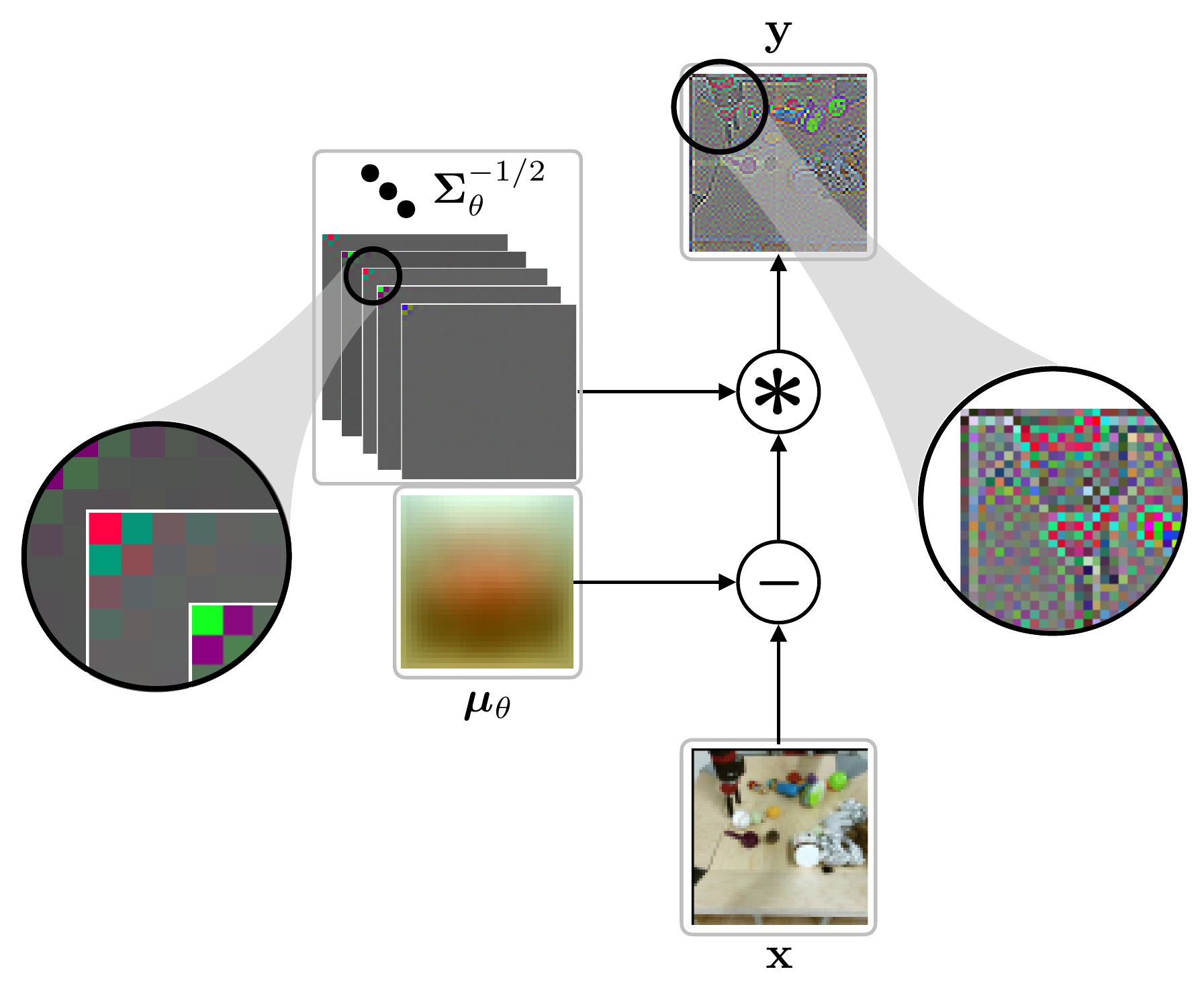}
    \caption{Spatial}
    \label{fig: pc spatial norm}
    \end{subfigure}% 
    ~
    \begin{subfigure}[t]{0.4\textwidth}
    \includegraphics[width=0.89\textwidth]{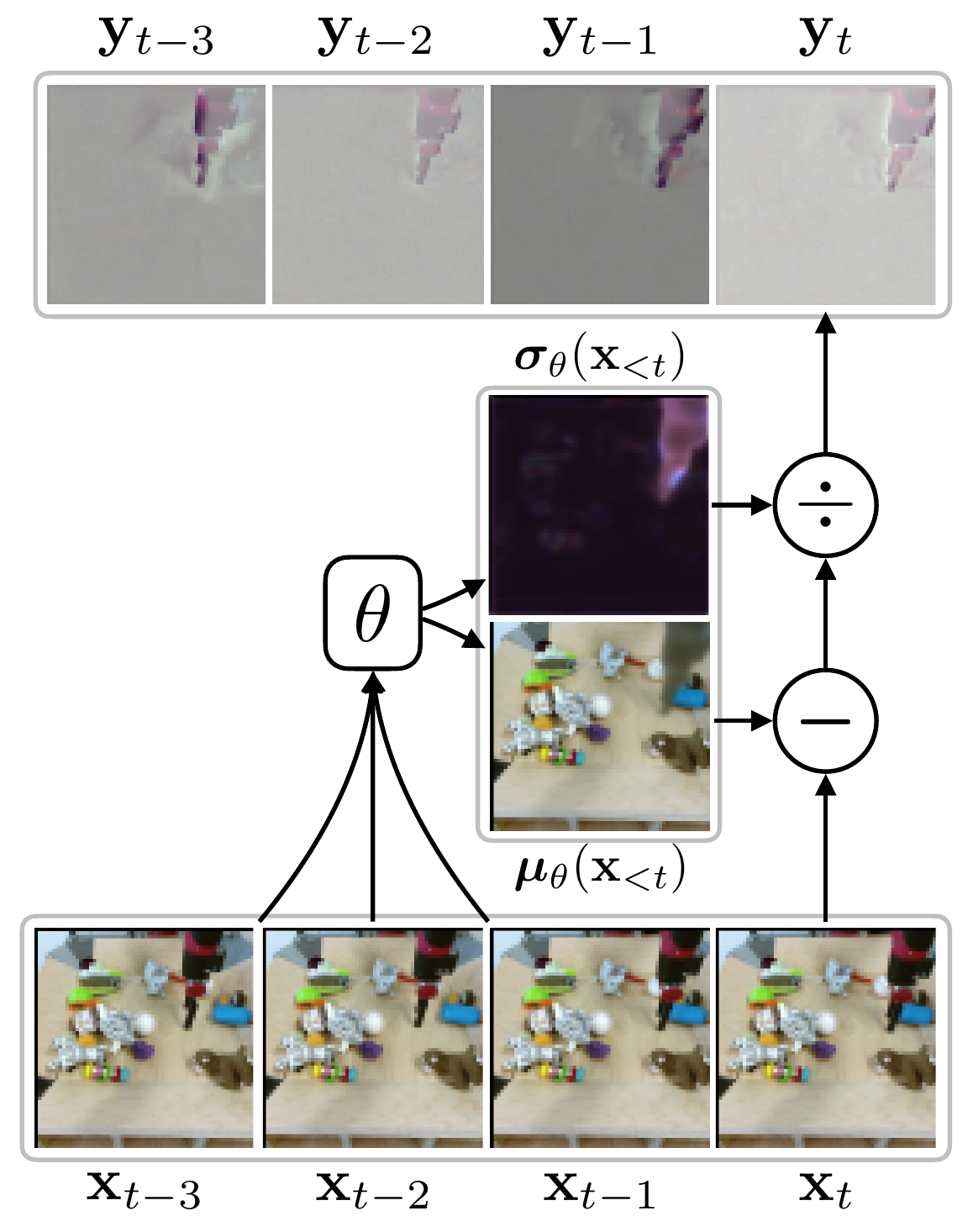}
    \caption{Temporal}
    \label{fig: pc temporal norm}
    \end{subfigure}
    \caption{\textbf{Spatiotemporal Predictive Coding}. \textbf{(a)} Spatial predictive coding removes spatial dependencies, using center-surround filters (left) to extract edges (right). \textbf{(b)} Temporal predictive coding removes temporal dependencies, extracting motion from video. Video frames are from BAIR Robot Pushing \citep{ebert2017self}.}
    \label{fig: pc spatiotemporal pc}
\end{figure}

Spatiotemporal predictive coding \citep{srinivasan1982predictive} forms predictions across spatial dimensions and temporal sequences. These predictions produce the resulting ``code'' as the prediction error. In the temporal setting, we can consider a Gaussian autoregressive model defined over observation sequences, $\mathbf{x}_{1:T}$. The conditional probability at time $t$ is
\begin{equation}
    p_\theta (\mathbf{x}_t | \mathbf{x}_{<t}) = \mathcal{N} (\mathbf{x}_t; \bm{\mu}_\theta (\mathbf{x}_{<t}), \textrm{diag} (\bm{\sigma}^2_\theta (\mathbf{x}_{<t}))). \nonumber
\end{equation}
Using auxiliary variables, $\mathbf{y}_t \sim \mathcal{N} (\mathbf{0}, \mathbf{I})$, we can express $\mathbf{x}_t = \bm{\mu}_\theta (\mathbf{x}_{<t}) + \bm{\sigma}_\theta (\mathbf{x}_{<t}) \odot \mathbf{y}_t$, where $\odot$ denotes element-wise multiplication. Conversely, we can express the inverse, normalization or \textit{whitening} transform as
\begin{equation}
    \mathbf{y}_t = \frac{\mathbf{x}_t - \bm{\mu}_\theta (\mathbf{x}_{<t})}{\bm{\sigma}_\theta (\mathbf{x}_{<t})},
    \label{eq: pc temporal normalization}
\end{equation}
which is a weighted prediction error.\footnote{\normalsize Note that other forms of probabilistic models will result in other forms of whitening transforms.} A video example is shown in Figure~\ref{fig: pc temporal norm}. The normalization transform removes temporal redundancy in the input, enabling the resulting sequence, $\mathbf{y}_{1:T}$, to be compressed more efficiently \citep{shannon1948mathematical, harrison1952experiments, oliver1952efficient}. This technique forms the basis of modern video \citep{wiegand2003overview} and audio \citep{atal1979predictive} compression. Note that one special case of this transform is $\bm{\mu}_\theta (\mathbf{x}_{<t}) = \mathbf{x}_{t-1}$ and $\bm{\sigma}_\theta (\mathbf{x}_{<t}) = \mathbf{1}$, in which case, $\mathbf{y}_t = \mathbf{x}_t - \mathbf{x}_{t-1} = \Delta \mathbf{x}_t$, i.e.,~temporal changes. For slowly changing sequences, this is a reasonable choice.

Normalization can also be applied within $\mathbf{x}_t$ to remove spatial dependencies. For instance, we can apply another autoregressive transform over spatial dimensions, predicting the $i^\textrm{th}$ dimension, $x_{i, t}$, as a function of previous dimensions, $\mathbf{x}_{1:i, t}$ (Eq.~\ref{eq: background ar non-temporal}). With linear functions, this corresponds to Cholesky whitening \citep{pourahmadi2011covariance, kingma2016improved}. However, this imposes an ordering over dimensions. Zero-phase components analysis (ZCA) whitening instead learns symmetric spatial dependencies \citep{kessy2018optimal}. Modeling these dependencies with a constant covariance matrix, $\bm{\Sigma}_\theta$, and mean, $\bm{\mu}_\theta$, the whitening transform is $\mathbf{y} = \bm{\Sigma}^{-1/2}_\theta (\mathbf{x} - \bm{\mu}_\theta)$. With natural images, this results in center-surround filters in the rows of $\bm{\Sigma}^{-1}_\theta$, thereby extracting edges (Figure~\ref{fig: pc spatial norm}).

\cite{srinivasan1982predictive} investigated spatiotemporal predictive coding in the retina, where compression is essential for transmission through the optic nerve. Estimating the (linear) auto-correlation of input sensory signals, they showed that spatiotemporal predictive coding models retinal ganglion cell responses in flies. This scheme allows these neurons to more fully utilize their dynamic range. It is generally accepted that retina, in part, performs stages of spatiotemporal normalization through center-surround receptive fields and on-off responses \citep{hosoya2005dynamic, graham2006can, pitkow2012decorrelation, palmer2015predictive}. \cite{dong1995temporal} applied similar ideas to thalamus, proposing an additional stage of temporal normalization. This also relates to the notion of generalized coordinates \citep{friston2008hierarchical}, i.e.,~modeling temporal derivatives, which can be approximated using finite differences (prediction errors). That is, $\frac{d \mathbf{x}}{dt} \approx \Delta \mathbf{x}_t \equiv \mathbf{x}_t - \mathbf{x}_{t-1}$. Thus, spatiotemporal predictive coding may be utilized at multiple stages of sensory processing to remove redundancy \citep{huang2011predictive}.

In neural circuits, normalization often involves inhibitory interneurons \citep{carandini2012normalization}, performing operations similar to those in Eq.~\ref{eq: pc temporal normalization}. For instance, inhibition occurs in retina between photoreceptors, via horizontal cells, and between bipolar cells, via amacrine cells. This can extract unpredicted motion, e.g.,~an object moving relative to the background \citep{olveczky2003segregation, baccus2008retinal}. A similar scheme is present in the lateral geniculate nucleus (LGN) in thalamus, with interneurons inhibiting relay cells from retina \citep{sherman2002role}. As mentioned above, this is thought to perform temporal normalization \citep{dong1995temporal, dan1996efficient}. Lateral inhibition is also prominent in neocortex, with distinct classes of interneurons shaping the responses of pyramidal neurons \citep{isaacson2011inhibition}. Part of their computational role appears to be spatiotemporal normalization \citep{carandini2012normalization}.
% Finally, while we have focused largely on early stages of sensory processing, inhibitory interneurons are also prevalent in other areas of neocortex, as well as in central pattern generator (CPG) circuits \citep{marder2001central}, found in the spinal cord. These circuits are repsonsible for the rhythmic generation of movement, such as locomotion. Thus, just as inhibitory interactions \textit{remove} spatiotemporal dependencies in early sensory areas, similar computational operations can \textit{add} spatiotemporal dependencies in motor activation.

\subsection{Hierarchical Predictive Coding}
\label{sec: pc hpc}

\begin{figure}[t]
\centering
\includegraphics[width=0.8\textwidth]{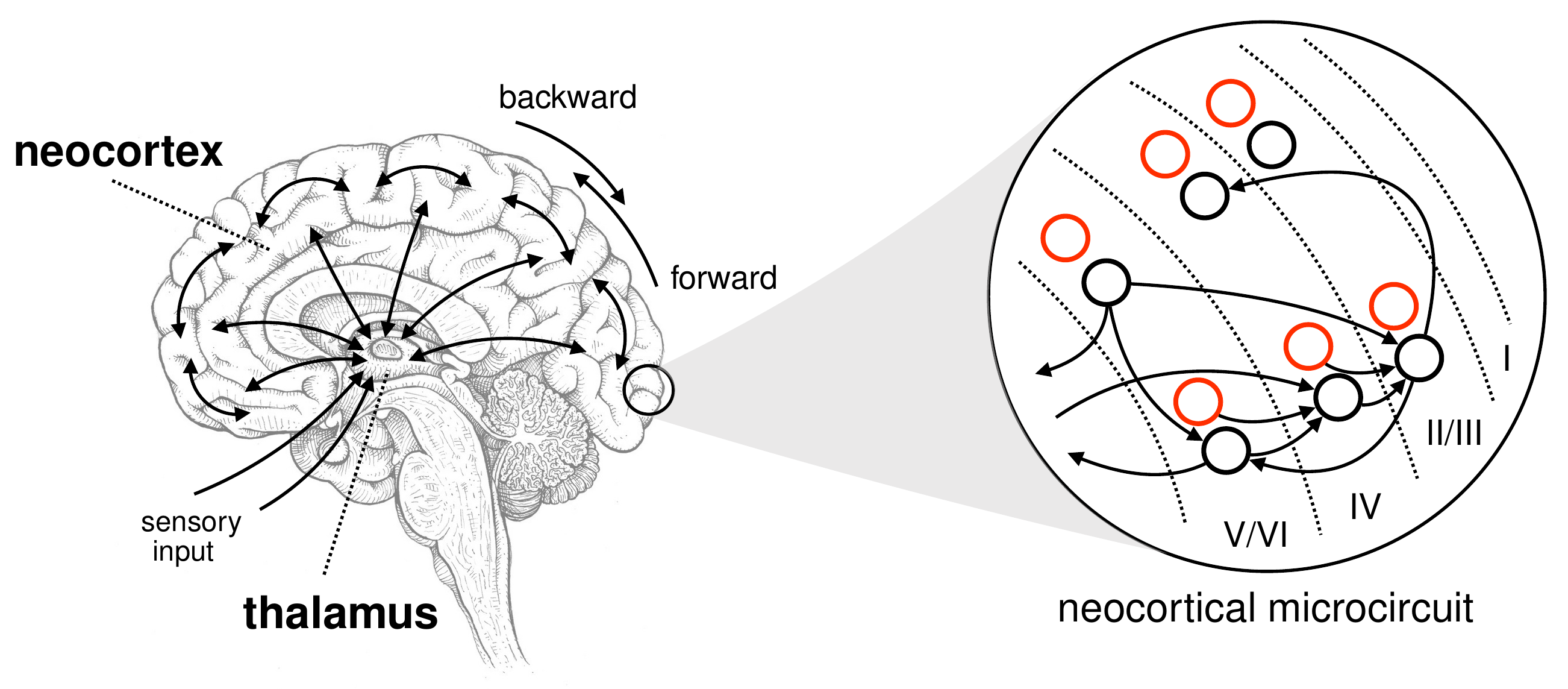}
\caption{\textbf{Brain Anatomy \& Cortical Circuitry}. Sensory inputs enter thalamus, forming reciprocal connections with neocortex. Neocortex is composed of six layers, with columns across layers and hierarchies of columns. Black and red circles represent excitatory and inhibitory neurons respectively, with arrows denoting connections. This circuit is repeated with variations throughout neocortex.}
\label{fig: brain anatomy}
\end{figure}

Hierarchical predictive coding has been postulated as a model of hierarchical processing in neocortex \citep{rao1999predictive, friston2005theory}, the outer sheet-like structure involved in sensory and motor processing (Figure~\ref{fig: brain anatomy}). Neocortex is composed of six layers (I--VI), with neurons arranged into columns, each engaged in related computations \citep{mountcastle1955topographic}. Columns interact locally via inhibitory interneurons, while also forming longer-range hierarchies via pyramidal neurons. Such hierarchies characterize multiple perceptual (and motor) processing pathways \citep{van1983hierarchical}. Longer-range connections are split into forward (up the hierarchy) and backward (down) directions. Forward connections are driving, evoking neural activity \citep{girard1989visual, girard1991visual}. Backward connections can be modulatory or driving \citep{covic2011synaptic, de2011synaptic}, which can be inverted through inhibition \citep{meyer2011inhibitory}. These connections, repeated with variations throughout neocortex, constitute a canonical \textit{neocortical microcircuit} \citep{douglas1989canonical}, suggesting a single algorithm \citep{hawkins2004intelligence}, capable of adapting to various inputs \citep{sharma2000induction}.

Formulating a theory of neocortex, \cite{mumford1992computational} described thalamus as an `active blackboard,' with the neocortex attempting to reconstruct the activity in thalamus and lower hierarchical areas. Under this theory, backward projections convey predictions, while forward projections use prediction errors to update estimates. Through a dynamic process, the system settles to an activity pattern, minimizing prediction error. Over time, the parameters are also adjusted to improve predictions. In this way, negative feedback is used, both in inference and learning, to construct a generative model of sensory inputs. Generative state estimation dates back (at least) to Helmholtz \citep{von1867handbuch}, and error-based updating is inline with cybernetics \citep{wiener1948cybernetics, mackay1956epistemological}, which influenced Kalman filtering \citep{kalman1960new}, a ubiquitous Bayesian filtering algorithm.

A mathematical formulation of Mumford's model, with ties to Kalman filtering (\cite{rao1998correlates}), was provided by \cite{rao1999predictive}, with the generalization to variational inference provided by \cite{friston2005theory}. To illustrate this setup, consider a model consisting of a single level of continuous latent variables, $\mathbf{z}$, modeling continuous data observations, $\mathbf{x}$. We will use Gaussian densities for each distribution and assume we have
\begin{align}
    p_\theta (\mathbf{x} | \mathbf{z}) & = \mathcal{N} (\mathbf{x}; f (\mathbf{W} \mathbf{z}), \textrm{diag} (\bm{\sigma}_\mathbf{x}^2)), \label{eq: MAP cond like} \\
    p_\theta (\mathbf{z} ) & = \mathcal{N} (\mathbf{z} ; \bm{\mu}_\mathbf{z}, \textrm{diag} (\bm{\sigma}_\mathbf{z}^2)), \label{eq: MAP prior}
\end{align}
where $f$ is an element-wise function (e.g.,~logistic sigmoid, tanh, or the identity), $\mathbf{W}$ is a weight matrix, $\bm{\mu}_\mathbf{z}$ is the prior mean, and $\bm{\sigma}^2_\mathbf{x}$ and $\bm{\sigma}^2_\mathbf{z}$ are vectors of variances.

In the simplest approach to inference, we can find the maximum-a-posteriori (MAP) estimate, i.e.,~estimate the $\mathbf{z}^*$ which maximizes $p_\theta (\mathbf{z} | \mathbf{x})$. While we cannot tractably evaluate $p_\theta (\mathbf{z} | \mathbf{x}) = \frac{p_\theta (\mathbf{x} , \mathbf{z})}{p_\theta (\mathbf{x})}$ directly, we can write
\begin{align}
    \mathbf{z}^* & = \textrm{arg} \max_{\mathbf{z}} \frac{p_\theta (\mathbf{x} , \mathbf{z})}{p_\theta (\mathbf{x})} \nonumber \\
    % & = \textrm{arg} \max_{\mathbf{z}} \frac{p_\theta (\mathbf{x} , \mathbf{z})}{p_\theta (\mathbf{x})} \nonumber \\
    & = \textrm{arg} \max_{\mathbf{z}} p_\theta (\mathbf{x} , \mathbf{z}). \nonumber
\end{align}
Thus, we can perform inference using the joint distribution, $p_\theta (\mathbf{x} , \mathbf{z}) = p_\theta (\mathbf{x} | \mathbf{z}) p_\theta (\mathbf{z})$, which we can tractably evaluate. We can also replace the optimization over the probability distribution with an optimization over the $\log$ probability, since $\log (\cdot)$ is a monotonically increasing function and does not affect the optimization. We then have
\begin{align}
    \mathbf{z}^* & = \textrm{arg} \max_{\mathbf{z}} \left[ \log p_\theta (\mathbf{x} | \mathbf{z} ) + \log p_\theta (\mathbf{z}) \right]. \nonumber \\
    & = \textrm{arg} \max_{\mathbf{z}} \left[ \log \mathcal{N} (\mathbf{x}; f (\mathbf{W} \mathbf{z}), \textrm{diag} (\bm{\sigma}_\mathbf{x}^2)) + \log \mathcal{N} (\mathbf{z} ; \bm{\mu}_\mathbf{z}, \textrm{diag} (\bm{\sigma}_\mathbf{z}^2)) \right]. \nonumber
    \label{eq: MAP inference 5}
\end{align}
Each of the terms in this objective is a weighted squared error. For instance, the first term is the weighted squared error in reconstructing the data observation:
\begin{equation}
\begin{split}
    \log \mathcal{N} (\mathbf{x}; f (\mathbf{W} \mathbf{z}), \textrm{diag} (\bm{\sigma}_\mathbf{x}^2)) = \frac{-M}{2} \log (2 \pi) -\frac{1}{2} \log \det \left( \textrm{diag}(\bm{\sigma}^2_\mathbf{x}) \right)  -\frac{1}{2} \left| \left| \frac{\mathbf{x} - f(\mathbf{W} \mathbf{z})}{\bm{\sigma}_\mathbf{x}} \right| \right|^2_2, \nonumber
    \label{eq: MAP recon error}
\end{split}
\end{equation}
where $M$ is the dimensionality of $\mathbf{x}$ and $|| \cdot ||^2_2$ denotes the squared L2 norm. Plugging these terms into the objective and dropping terms that do not depend on $\mathbf{z}$ yields
\begin{equation}
    \mathbf{z}^* = \textrm{arg} \max_{\mathbf{z}} \underbrace{ \left[ \frac{-1}{2} \left| \left| \frac{\mathbf{x} - f(\mathbf{W} \mathbf{z})}{\bm{\sigma}_\mathbf{x}} \right| \right|^2_2  - \frac{1}{2} \left| \left| \frac{\mathbf{z} - \bm{\mu}_\mathbf{z}}{\bm{\sigma}_\mathbf{z}} \right| \right|^2_2 \right] }_{\mathscr{L} (\mathbf{z}; \theta)}, \label{eq: MAP log objective 3}
    % & = \textrm{arg} \max_{\mathbf{z}} \mathscr{L} (\mathbf{z}; \theta), \nonumber
\end{equation}
where we have defined the objective as $\mathscr{L} (\mathbf{z}; \theta)$. For purposes of illustration, let us assume that $f (\cdot)$ is the identity function, i.e.,~$f(\mathbf{W}\mathbf{z}) = \mathbf{W}\mathbf{z}$. We can then evaluate the gradient of $\mathscr{L} (\mathbf{z}; \theta)$ w.r.t.~$\mathbf{z}$, yielding
\begin{equation}
    \nabla_\mathbf{z} \mathscr{L} (\mathbf{z}; \theta) = \mathbf{W}^\intercal \underbrace{ \left( \frac{\mathbf{x} - \mathbf{W} \mathbf{z}}{\bm{\sigma}_\mathbf{x}} \right)}_{ \bm{\xi}_\mathbf{x}} - \underbrace{ \frac{\mathbf{z} - \bm{\mu}_\mathbf{z}}{\bm{\sigma}_\mathbf{z}}}_{ \bm{\xi}_\mathbf{z}}.
    \label{eq: MAP inference gradient}
\end{equation}
\begin{figure}[t!]
    \centering
    \includegraphics[width=0.9\textwidth]{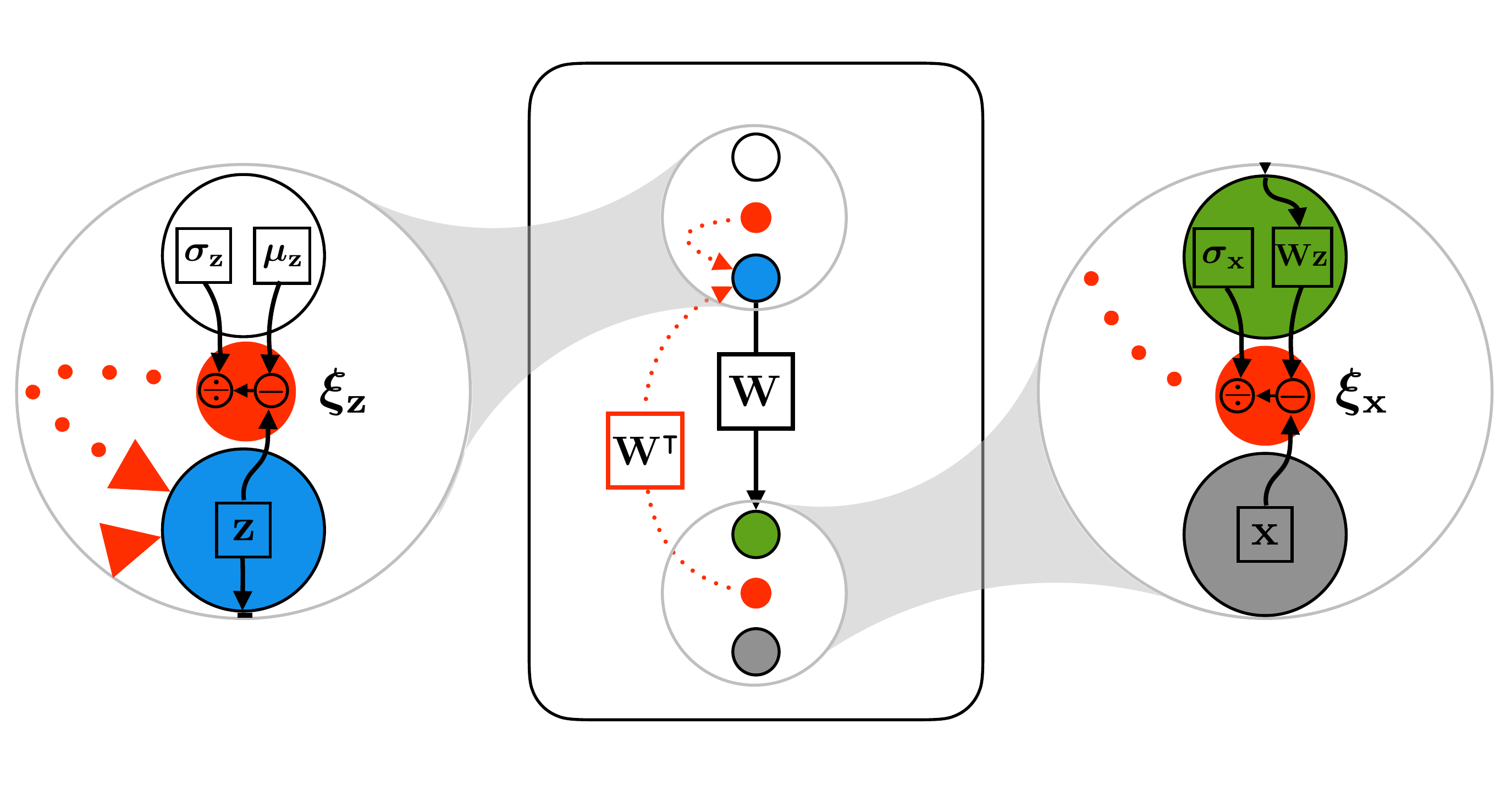}
    \caption{\textbf{Hierarchical Predictive Coding}. The diagram shows the basic computation graph for a Gaussian latent variable model with MAP inference. The insets show the weighted error calculation for the latent (left) and observed (right) variables.}
    \label{fig: pc hierarchical pc}
\end{figure}
The transposed weight matrix, $\mathbf{W}^\intercal$, results from differentiating $\mathbf{W}\mathbf{z}$, translating the reconstruction error into an update in $\mathbf{z}$. We have also defined the weighted errors, $\bm{\xi}_\mathbf{x}$ and $\bm{\xi}_\mathbf{z}$. From Eq.~\ref{eq: MAP inference gradient}, we see that if we want to perform inference using gradient-based optimization, e.g., $\mathbf{z} \leftarrow \mathbf{z} + \alpha \nabla_\mathbf{z} \mathscr{L} (\mathbf{z}; \theta)$, we need 1) the weighted errors, $\bm{\xi}_\mathbf{x}$ and $\bm{\xi}_\mathbf{z}$, and 2) the transposed weights, $\mathbf{W}^\intercal$, or more generally, the Jacobian of the conditional likelihood mean. This overall scheme is depicted in Figure~\ref{fig: pc hierarchical pc}.

To learn the weight parameters, we can differentiate $\mathscr{L} (\mathbf{z}; \theta)$ (Eq.~\ref{eq: MAP log objective 3}) w.r.t.~$\mathbf{W}$:
\begin{align}
    \nabla_\mathbf{W} \mathscr{L} (\mathbf{z}; \theta) & = \bm{\xi}_\mathbf{x} \mathbf{z}^\intercal . \nonumber
    \label{eq: MAP weight gradient}
\end{align}
This gradient is the product of a local error, $\bm{\xi}_\mathbf{x}$, and the latent variable, $\mathbf{z}$, suggesting the possibility of a biologically-plausible learning rule \citep{whittington2017approximation}.

Predictive coding identifies the conditional likelihood (Eq.~\ref{eq: MAP cond like}) with backward (top-down) cortical projections, whereas inference (Eq.~\ref{eq: MAP inference gradient}) is identified with forward (bottom-up) projections \citep{friston2005theory}. Each are thought to be mediated by pyramidal neurons. Under this model, each cortical column predicts and estimates a stochastic continuous latent variable, possibly represented via a (pyramidal) firing rate or membrane potential \citep{friston2005theory}. Interneurons within columns calculate errors ($\bm{\xi}_\mathbf{x}$ and $\bm{\xi}_\mathbf{z}$). Although we have only discussed diagonal covariance ($\bm{\sigma}^2_\mathbf{x}$ and $\bm{\sigma}^2_\mathbf{z}$), lateral inhibitory interneurons could parameterize full covariance matrices, i.e.,~$\bm{\Sigma}_\mathbf{x}$ and $\bm{\Sigma}_\mathbf{z}$, as a form of \textit{spatial} predictive coding (Section \ref{sec: pc spatiotemporal}). These factors weight $\bm{\xi}_\mathbf{x}$ and $\bm{\xi}_\mathbf{z}$, modulating the \textit{gain} of each error as a form of ``attention'' \citep{feldman2010attention}. Neural correspondences are summarized in Table~\ref{table: biological correspondences}.

We have presented a simplified model of hierarchical predictive coding, without multiple latent levels and dynamics. A full hierarchical predictive coding model would include these aspects and others. In particular, Friston has explored various design choices \citep{friston2007variational, friston2008hierarchical, friston2008variational}, yet the core aspects of probabilistic generative modeling and variational inference remain the same. Elaborating and comparing these choices will be essential for empirically validating hierarchical predictive coding.

\begin{table}[t!]
\centering
\caption{\textbf{Neural Correspondences of Hierarchical Predictive Coding}.}
\begin{tabular}{c|c}
\toprule
\textbf{Neuroscience}  & \textbf{Predictive Coding} \\  \bottomrule
Top-Down Cortical Projections & Generative Model Conditional Mapping  \\ \hline
Bottom-Up Cortical Projections & Inference Updating \\ \hline
Lateral Inhibition & Covariance Matrices  \\ \hline
(Pyramidal) Neuron Activity & Latent Variable Estimates \& Errors \\ \hline
Cortical Column & Corresponding Estimate \& Error \\
\bottomrule
\end{tabular}
\label{table: biological correspondences}
\end{table}

\subsection{Empirical Support}

While there is considerable evidence in support of predictions and errors in neural systems, disentangling these general aspects of predictive coding from the particular algorithmic choices remains challenging \citep{gershman2019does}. Here, we outline relevant work, but we refer to \citet{huang2011predictive, bastos2012canonical, clark2013whatever, keller2018predictive, walsh2020evaluating} for a more in-depth overview.

\paragraph{Spatiotemporal} Various works have investigated predictive coding in early sensory areas, e.g., retina \citep{srinivasan1982predictive, atick1992does}. This involves fitting retinal ganglion cell responses to a spatial whitening (or decorrelation) process \citep{graham2006can, pitkow2012decorrelation}, which can be dynamically adjusted \citep{hosoya2005dynamic}. Similar analyses suggest that retina also employs temporal predictive coding \citep{srinivasan1982predictive, palmer2015predictive}. Such models typically contain linear whitening filters (center-surround) followed by non-linearities. These non-linearities have been shown to be essential for modeling responses \citep{pitkow2012decorrelation}, possibly by inducing added sparsity \citep{graham2006can}. Spatiotemporal predictive coding also appears to be found in thalamus \citep{dong1995temporal, dan1996efficient} and cortex, however, such analyses are complicated by backward, modulatory inputs.

\paragraph{Hierarchical} Early work toward empirically validating hierarchical predictive coding came from explaining extra-classical receptive field effects \citep{rao1999predictive, rao2002predictive}, whereby top-down signals in cortex alter classical visual receptive fields, suggesting that top-down influences play a key role in sensory processing \citep{gilbert2007brain}. Note that such effects support a cortical generative model generally \citep{olshausen1997sparse}, not predictive coding specifically.

Temporal influences have been demonstrated through repetition suppression \citep{summerfield2006predictive}, in which activity diminishes in response to repeated, i.e.,~predictable, stimuli. This may reflect error suppression from improved predictions. Predictive coding has also been used to explain biphasic responses in LGN \citep{jehee2009predictive}, in which reversing the visual input with an anti-correlated image results in a large response, presumably due to prediction errors. Predictive signals have been documented in auditory \citep{wacongne2011evidence} and visual \citep{meyer2011statistical} processing. Activity seemingly corresponding to prediction errors has also been observed in a variety of areas and contexts, including visual cortex in mice \citep{keller2012sensorimotor, zmarz2016mismatch, gillon2021learning}, auditory cortex in monkeys \citep{eliades2008neural} and rodents \citep{parras2017neurons}, and visual cortex in humans \citep{murray2002shape, alink2010stimulus, egner2010expectation}. Thus, sensory cortex appears to be engaged in hierarchical and temporal prediction, with prediction errors playing a key role.

Empirical evidence for predictive coding aside, given the complexity of neural systems, the theory is undoubtedly incomplete or incorrect. Without the low-level details, e.g., connectivity, potentials, etc., it is difficult to determine the computational form of the circuit. Further, these models are typically oversimplified, with few trained parameters, detached from natural stimuli. While new tools enable us to test predictive coding in neural circuits \citep{gillon2021learning}, machine learning, particularly VAEs, can advance from the other direction. Training large-scale models on natural stimuli may improve empirical predictions for biological systems  \citep{rao1999predictive, lotter2018neural}.

% Thus, while general aspects of predictive coding are supported, we are unable to probe into the details of such models, making predictive coding a largely \textit{normative} theory. The purpose of this paper is to establish connections between predictive coding and machine learning. Ideally, by building larger-scale models and training them on similar sensory data, we can form more fine-grained empirical predictions for biological neural systems. Building off of the example of \cite{rao1999predictive}, \cite{lotter2018neural} provided another step in this direction, comparing the responses of neural systems and their hierarchical predictive coding model. In this paper, we have attempted to help further build the foundation for this collaborative effort.

\section{Variational Autoencoders}
\label{sec: vaes}

Variational autoencoders (VAEs) \citep{kingma2014stochastic, rezende2014stochastic} are latent variable models parameterized by deep networks. As in hierarchical predictive coding, these models typically contain Gaussian latent variables and are trained using variational inference. However, rather than performing inference optimization directly, VAEs \textit{amortize} inference \citep{gershman2014amortized}.

\subsection{Amortized Variational Inference}

Amortization refers to spreading out costs. In amortized inference, these are the computational costs of inference. With $q (\mathbf{z} | \mathbf{x}) = \mathcal{N} (\mathbf{z}; \bm{\mu}_q , \diag ( \bm{\sigma}^2_q ))$ and $\bm{\lambda} \equiv \left[ \bm{\mu}_q , \bm{\sigma}_q  \right]$, rather than separately optimizing $\bm{\lambda}$ for each data example, we amortize this optimization using a learned optimizer or \textit{inference model}. By using meta-optimization, we can perform inference optimization far more efficiently. Inference models are linked with deep latent variable models, popularized by the Helmholtz Machine \citep{dayan1995helmholtz}, a form of \textit{autoencoder} \citep{ballard1987modular}. Here, the inference model is a direct mapping from $\mathbf{x}$ to $\bm{\lambda}$:
\begin{equation}
    \bm{\lambda} \leftarrow f_\phi (\mathbf{x}),
    \label{eq: background direct amortization}
\end{equation}
where $f_\phi$ is a function (deep network) with parameters $\phi$. We then denote the approximate posterior as $q_\phi (\mathbf{z} | \mathbf{x})$ to denote the parameterization by $\phi$. Now, rather than optimizing $\bm{\lambda}$ using gradient-based techniques, we update $\phi$ using $\nabla_\phi \mathcal{L} = \frac{\partial \mathcal{L}}{\partial \bm{\lambda}} \frac{\partial \bm{\lambda}}{\partial \phi}$, thereby letting $f_\phi$ learn to optimize $\bm{\lambda}$. This procedure is simple, as we only need to tune the learning rate for $\phi$, and efficient, as we have an estimate of $\bm{\lambda}$ after one forward pass through $f_\phi$. Amortization is also widely applicable: if we can estimate $\nabla_{\bm{\lambda}} \mathcal{L}$, we can continue differentiating through the chain $\phi \rightarrow \bm{\lambda} \rightarrow \mathbf{z} \rightarrow \mathcal{L}$.

To differentiate through $\mathbf{z} \sim q_\phi (\mathbf{z} | \mathbf{x})$, we can use the pathwise derivative estimator, also referred to as the reparameterization estimator \citep{kingma2014stochastic}. This is accomplished by expressing $\mathbf{z}$ in terms of an auxiliary random variable. The most common example expresses $\mathbf{z} \sim \mathcal{N} (\mathbf{z}; \bm{\mu}_q, \diag ( \bm{\sigma}_q^2 ))$ as $\mathbf{z} = \bm{\mu}_q + \bm{\epsilon} \odot \bm{\sigma}_q$, where $\bm{\epsilon} \sim \mathcal{N} (\bm{\epsilon}; \mathbf{0}, \mathbf{I})$ and $\odot$ denotes element-wise multiplication. We can then estimate $\nabla_{\bm{\mu}_q} \mathcal{L}$ and $\nabla_{\bm{\sigma}_q} \mathcal{L}$, allowing us to calculate the inference model gradients, $\nabla_\phi \mathcal{L}$.

\begin{figure}[t]
    \centering
    \includegraphics[width=0.7\textwidth]{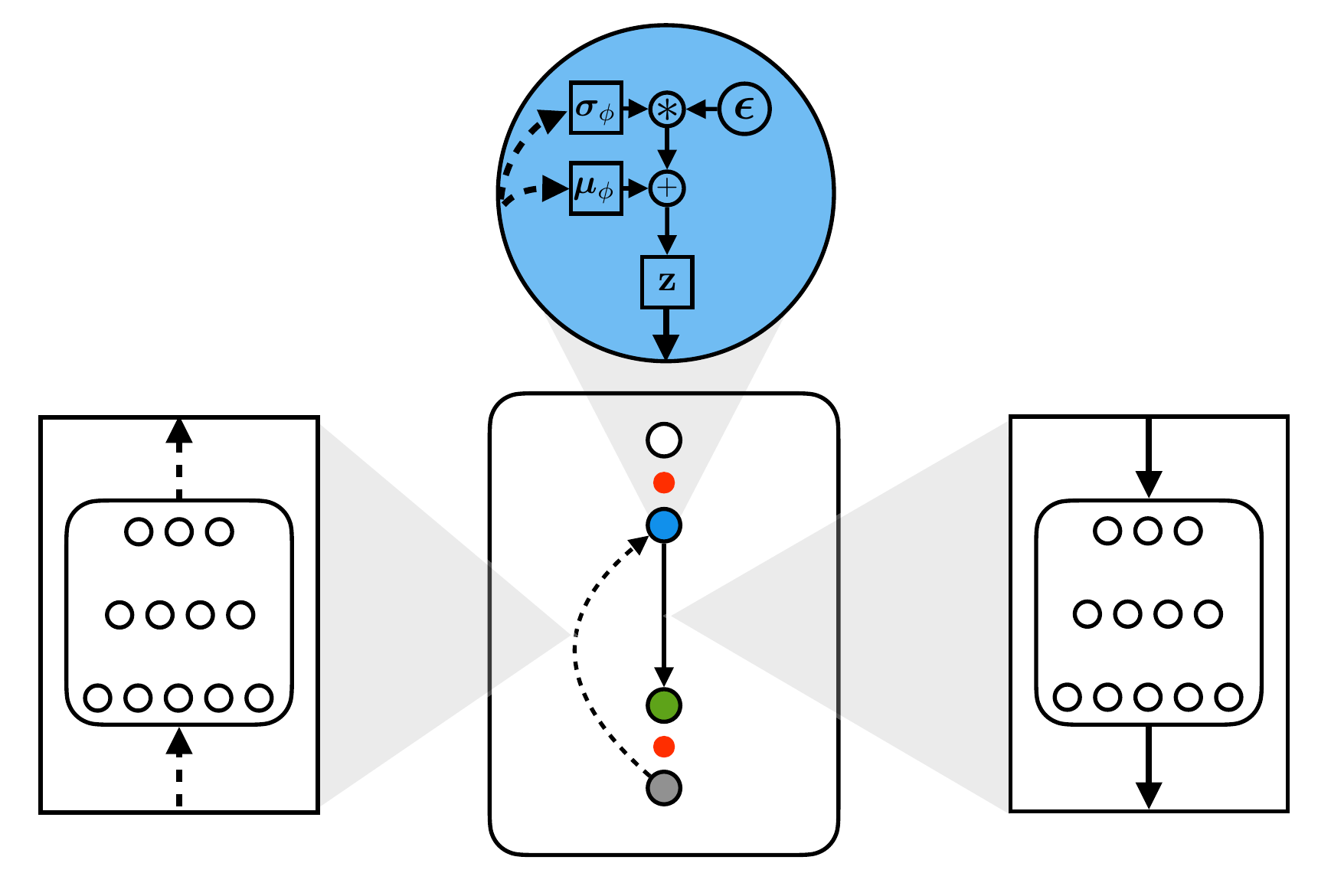}
    \caption{\textbf{Variational Autoencoder}. VAEs use direct amortization (Eq.~\ref{eq: background direct amortization}) to train deep latent variable models. The inference model (left) is an \textit{encoder}, and the conditional likelihood (right) is a \textit{decoder}. Each are parameterized by deep networks.}
    \label{fig: background vae comp graph}
\end{figure}

When direct amortization is combined with the pathwise derivative estimator in deep latent variable models, the resulting setup is a \textit{variational autoencoder} \citep{kingma2014stochastic, rezende2014stochastic}. In this interpretation, $q_\phi (\mathbf{z} | \mathbf{x})$ is an \textit{encoder}, $\mathbf{z}$ is the latent \textit{code}, and $p_\theta (\mathbf{x} | \mathbf{z})$ is a \textit{decoder} (Figure~\ref{fig: background vae comp graph}). This direct encoding scheme is intuitive: in the same way that $p_\theta (\mathbf{x} | \mathbf{z})$ directly maps $\mathbf{z}$ to a distribution over $\mathbf{x}$, $q_\phi (\mathbf{z} | \mathbf{x})$ directly maps $\mathbf{x}$ to a distribution over $\mathbf{z}$. Indeed, with perfect knowledge of $p_\theta (\mathbf{x}, \mathbf{z})$, $f_\phi$ could act as a lookup table, mapping each $\mathbf{x}$ to the corresponding optimal $\bm{\lambda}$. However, in practice, direct amortization of this form tends to result in suboptimal estimates of $\bm{\lambda}$ \citep{cremer2018inference}, motivating more powerful amortized inference techniques. 

\subsubsection{Iterative Amortized Inference}

One method for improving direct amortization involves incorporating iterative updates \citep{hjelm2016iterative, krishnan2017challenges, kim2018semi, marino2018iterative}, replacing a one-step inference procedure with a multi-step procedure. Iterative amortized inference \citep{marino2018iterative} maintains an inference model, but uses it to perform iterative updates on the approximate posterior. Following the previous notation, the basic form of an iterative amortized inference model is given as:
\begin{equation}
    \bm{\lambda} \leftarrow f_\phi (\bm{\lambda}, \nabla_{\bm{\lambda}} \mathcal{L}).
    \label{eq: iterative inference form}
\end{equation}
Iterative inference models take in the current estimate, $\bm{\lambda}$, as well as the gradient, $\nabla_{\bm{\lambda}} \mathcal{L}$, and output an updated estimate of $\bm{\lambda}$. As before, inference model parameters are updated using estimates of $\nabla_\phi \mathcal{L}$. Note that Eq.~\ref{eq: iterative inference form} generalizes stochastic gradient-based optimization. For instance, a special case is $\bm{\lambda} \leftarrow \bm{\lambda} + \alpha \nabla_{\bm{\lambda}} \mathcal{L}$, where $\alpha$ is a step-size, however, Eq.~\ref{eq: iterative inference form} also includes non-linear updates \citep{andrychowicz2016learning}.

In latent Gaussian models, $\nabla_{\bm{\lambda}}$ is defined by the weighted errors, $\bm{\xi}_\mathbf{x}$ and $\bm{\xi}_\mathbf{z}$, and the Jacobian of the conditional likelihood, $\mathbf{J}$ ($\mathbf{W}$ in the linear model in Section~\ref{sec: pc hpc}). Thus, in latent Gaussian models, we can consider inference models of the special form:
\begin{equation}
    \bm{\lambda} \leftarrow f_\phi (\bm{\lambda}, \bm{\xi}_\mathbf{x}, \bm{\xi}_\mathbf{z}).
    \label{eq: error-encoding model}
\end{equation}
This is a learned, non-linear mapping from errors to updated estimates of the approximate posterior, i.e.,~learned negative feedback. The distinction between direct and iterative amortization is shown in Figures~\ref{fig: pc vae diagram}~\&~\ref{fig: pc it am diagram}. Iterative amortization can be readily extended to sequential models \citep{marino2018general}, resulting in a general predict-update inference scheme, highly reminiscent of Kalman filtering \citep{kalman1960new}.

% \cite{marino2020iterative} also applied iterative amortized inference to perform policy optimization in reinforcement learning, yielding a form of learned feedback control.

\subsection{Extensions of VAEs}

\subsubsection{Additional Dependencies \& Representation Learning}

VAEs have been extended to a variety of architectures, incorporating hierarchical and temporal dependencies. \cite{sonderby2016ladder} proposed a hierarchical VAE, in which the conditional prior at each level, $\ell$, i.e., $p_\theta (\mathbf{z}^\ell | \mathbf{z}^{\ell+1 : L})$, is parameterized by a deep network (see Eq.~\ref{eq: background hierarchical lvm}). Follow-up works have scaled this approach with impressive results \citep{kingma2016improved, vahdat2020nvae, child2020very}, extracting increasingly abstract features at higher levels \citep{maaloe2019biva}. Another line of work has incorporated temporal dependencies within VAEs, parameterizing dynamics in the prior and conditional likelihood with deep networks \citep{chung2015recurrent, fraccaro2016sequential}. Such models can also provide representations and predictions for reinforcement learning \citep{ha2018recurrent, hafner2019learning}.

Other works have investigated representation learning within VAEs. One approach, the $\beta$-VAE \citep{higgins2017beta}, modifies the ELBO (Eq.~\ref{eq: elbo def 2}) by adjusting a weighting, $\beta$, on $D_\textrm{KL} (q (\mathbf{z} | \mathbf{x}) || p_\theta (\mathbf{z}))$. This tends to yield more disentangled, i.e., independent, latent variables. Indeed, $\beta$ controls the rate-distortion trade-off between latent complexity and reconstruction \citep{alemi2018fixing}, highlighting VAEs' ability to extract latent structure at multiple resolutions \citep{rezende2018taming}. Likewise, a separate line of work has focused on identifiability, i.e., the ability to uniquely recover the original latent variables within a model (or their posterior). While this is true in linear ICA \citep{comon1994independent}, it is not generally the case with non-linear ICA and non-invertible models (VAEs) \citep{khemakhem2020variational, gresele2020relative}, requiring special considerations.

\subsubsection{Normalizing Flows}
\label{sec: pc normalizing flows}

Another direction within VAEs is the use of normalizing flows \citep{rezende2015variational}. Flow-based distributions use invertible transforms to add and remove dependencies (Section~\ref{sec: background dep struct}). While such models can operate as generative models \citep{dinh2015nice, dinh2017density, papamakarios2017masked}, they can also define distributions in VAEs. This includes the approximate posterior \citep{rezende2015variational}, prior \citep{huang2017learnable}, and conditional likelihood \citep{agrawal2016deep}. In each case, a deep network outputs the parameters (e.g.,~mean and variance) of a \textit{base distribution} over a normalized variable. Separate deep networks parameterize the transforms, which map between the normalized and unnormalized variables.

\begin{figure}[t!]
    \centering
    \includegraphics[width=0.75\textwidth]{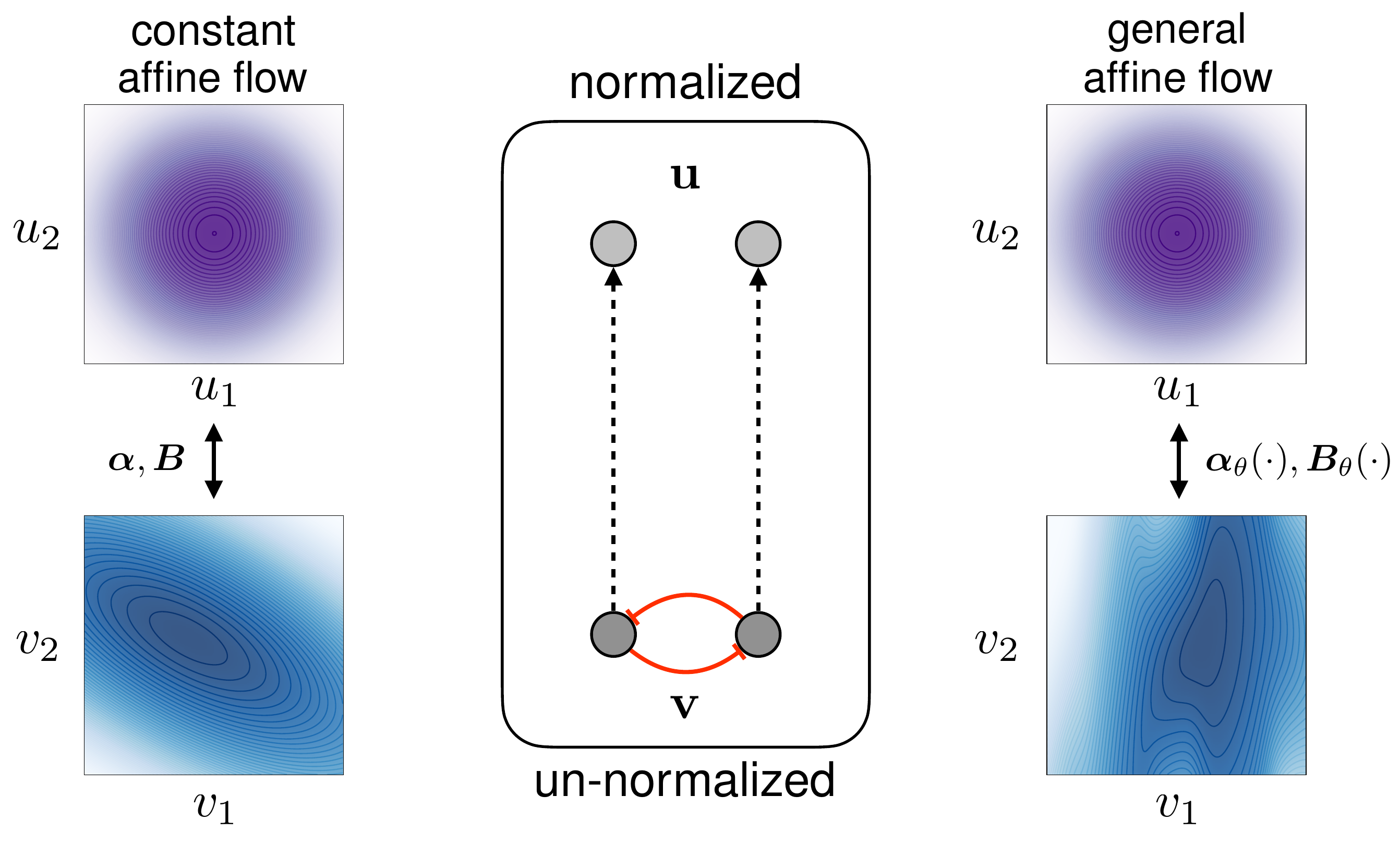}
    \caption{\textbf{Normalizing Flows}. Normalizing flows is a framework for adding or removing dependencies. Using affine parameter functions, $\bm{\alpha}_\theta$ and $\bm{B}_\theta$, one can model non-linear dependencies, generalizing constant transforms, e.g., a covariance matrix.}
    \label{fig: pc add remove dep}
\end{figure}

\paragraph{Example:} Consider a normalized variable, $\mathbf{u}$, defined by the distribution $p_\theta (\mathbf{u} | \cdot) = \mathcal{N} (\mathbf{u}; \bm{\mu}_\theta (\cdot), \textrm{diag} (\bm{\sigma}_\theta^2 (\cdot)))$, where $\bm{\mu}_\theta$ and $\bm{\sigma}_\theta$ are output by deep networks, with $\cdot$ denoting conditioning input variables. We consider an \textit{affine} transform \citep{dinh2017density}, defined by a shift vector, $\bm{\alpha} = \bm{\alpha}_\theta (\mathbf{u})$, and a scale matrix, $\bm{B} = \bm{B}_\theta (\mathbf{u})$, each of which may be parameterized by deep networks. This defines a new, un-normalized variable $\mathbf{v}$:
\begin{equation}
    \mathbf{v} = \bm{\alpha} + \bm{B} \mathbf{u},
    \label{eq: nf affine}
\end{equation}
which can now contain affine dependencies between dimensions. For Eq.~\ref{eq: nf affine} to be invertible, we require that $\bm{B}$ itself is invertible, i.e.,~non-zero determinant. Thus, $\bm{B}$ is a square matrix, and $\mathbf{u}$ and $\mathbf{v}$ are the same dimensionality. If, instead, we are given $\mathbf{v}$, we can calculate its log-probability by applying the normalizing inverse transform to get $\mathbf{u}$:
\begin{equation}
    \mathbf{u} = \bm{B}^{-1}(\mathbf{v} - \bm{\alpha}),
\end{equation}
then use the change of variables formula (Eq.~\ref{eq: background change of variables}). This converts the log-probability calculation from the structured space of $\mathbf{v}$ to the normalized space of $\mathbf{u}$.
% However, we need to account for the local scaling of space induced by $\bm{B}$ through its determinant.
Note that the multivariate Gaussian density (Eq.~\ref{eq: gaussian def}) is a special case of this transform, taking a standard Gaussian variable, $\mathbf{u} \sim \mathcal{N} (\mathbf{u}; \mathbf{0}, \mathbf{I})$, and adding linear dependencies to yield a multivariate Gaussian variable, $\mathbf{v} \sim \mathcal{N} (\mathbf{v}; \bm{\alpha}, \bm{B}^\intercal \bm{B})$. In this case, where $\bm{\alpha}$ and $\bm{B}$ are constant, the inverse transform removes linear dependencies between dimensions in $\mathbf{v}$. We depict this scheme in Figure \ref{fig: pc add remove dep}, along with the more general non-linear version provided by normalizing flows, in which $\bm{\alpha}_\theta$ and $\bm{B}_\theta$ are functions. Thus, normalizing flows provides a powerful, more general approach for augmenting the distributions in latent variable models, applicable both across space \citep{rezende2015variational, kingma2016improved} and time \citep{van2018parallel, marino2020improving}.

\section{Connections \& Comparisons}
\label{sec: pc connections}

\begin{figure}[t!]
    \centering
    \begin{subfigure}[t]{0.32\textwidth}
    \centering
    \includegraphics[width=0.75\textwidth]{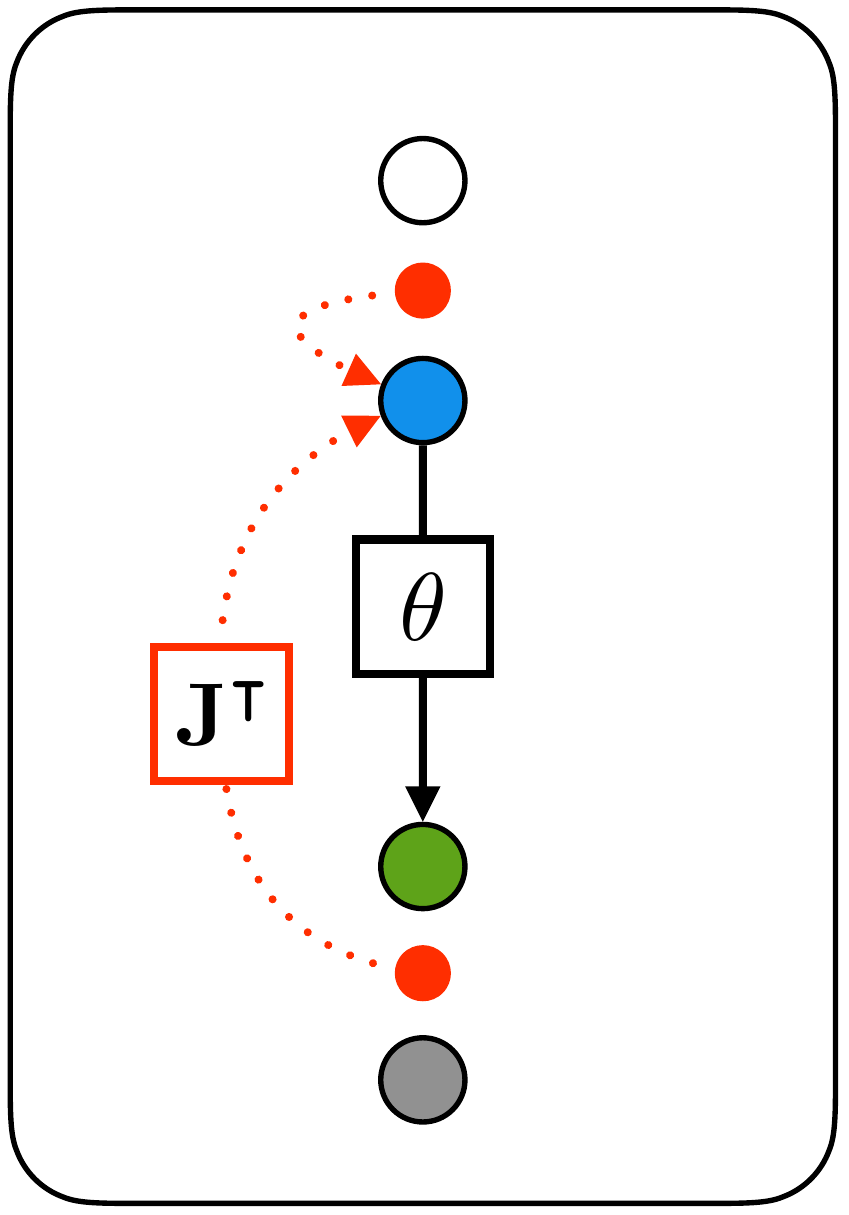}
    \caption{Predictive Coding}
    \label{fig: pc hierarchical pc diagram}
    \end{subfigure}% 
    ~
    \begin{subfigure}[t]{0.32\textwidth}
    \centering
    \includegraphics[width=0.75\textwidth]{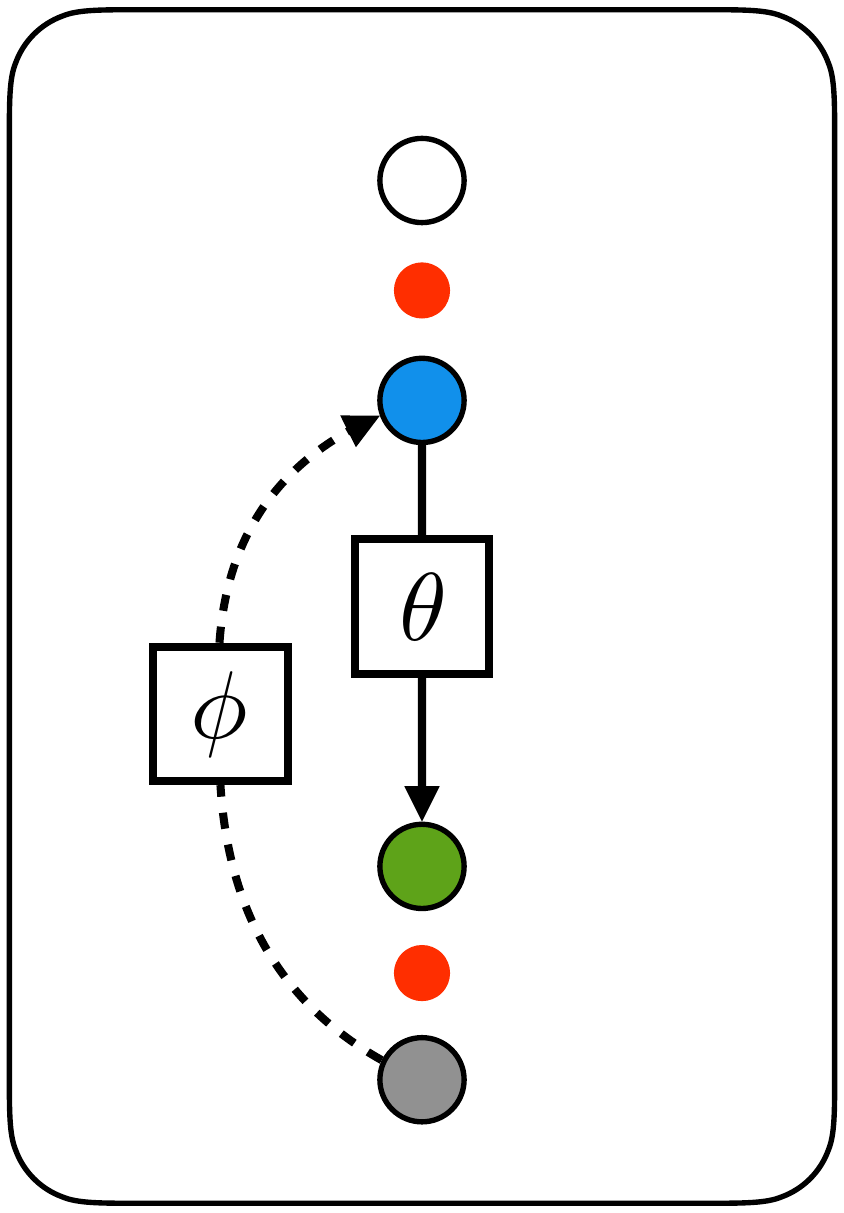}
    \caption{VAE (Direct)}
    \label{fig: pc vae diagram}
    \end{subfigure}% 
    ~
    \begin{subfigure}[t]{0.32\textwidth}
    \centering
    \includegraphics[width=0.75\textwidth]{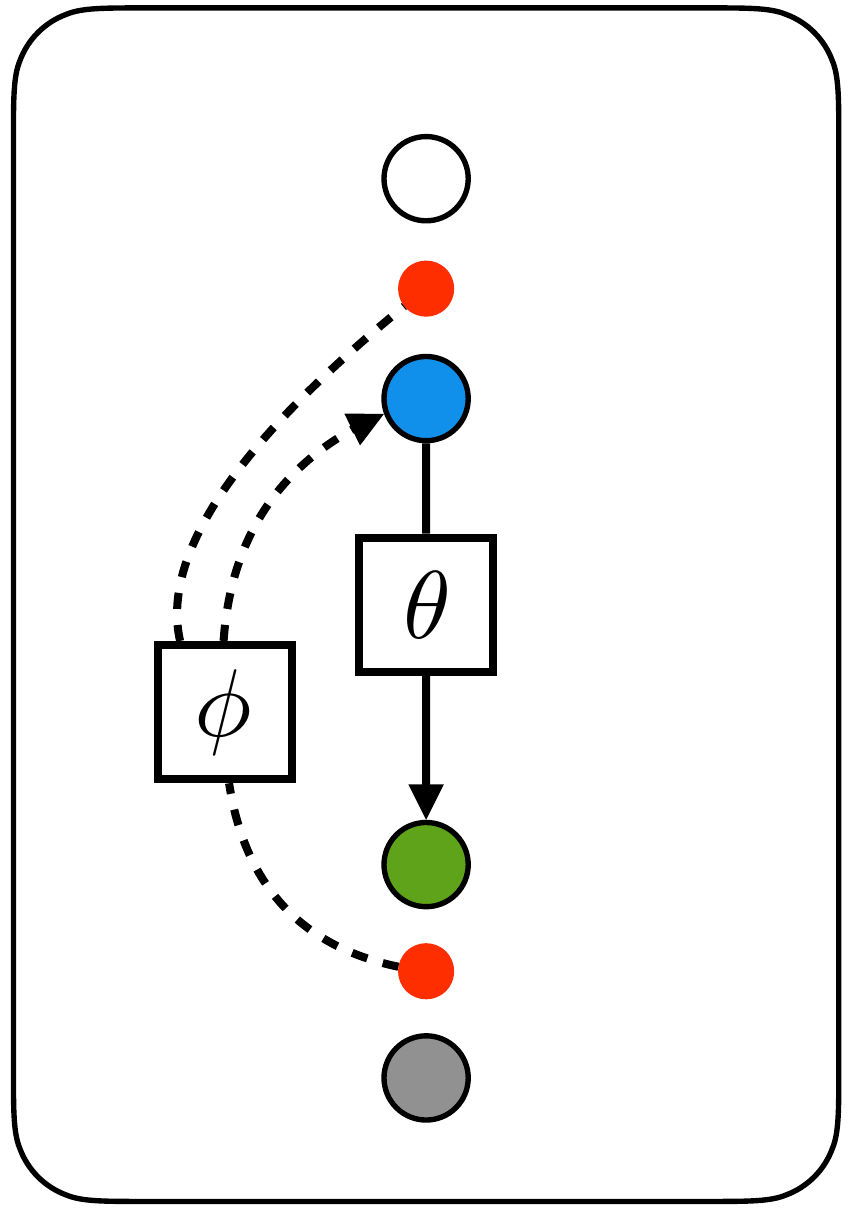}
    \caption{VAE (Iterative)}
    \label{fig: pc it am diagram}
    \end{subfigure}
    \caption{\textbf{Hierarchical Predictive Coding \& VAEs}. Computation diagrams for \textbf{(a)} hierarchical predictive coding, \textbf{(b)} VAE with direct amortized inference, and \textbf{(c)} VAE with iterative amortized inference \citep{marino2018iterative}. $\mathbf{J}^\intercal$ denotes the transposed Jacobian matrix of the conditional likelihood. Red dotted lines denote gradients, and black dashed lines denote amortized inference.}
    \label{fig: pc pc vae comparison}
\end{figure}

Predictive coding and VAEs (and deep generative models generally), are highly related, both in their model formulations and inference approaches (Figure~\ref{fig: pc pc vae comparison}). Specifically,
\begin{itemize}
    \item \textbf{Model Formulation}: Both areas consider hierarchical latent Gaussian models with non-linear dependencies between latent levels, as well as dependencies within levels via covariance matrices (predictive coding) or normalizing flows (VAEs).
    \item \textbf{Inference}: Both areas use variational inference, often with Gaussian approximate posteriors. While predictive coding and VAEs employ differing optimization techniques, these are design choices in solving the same inference problem.
\end{itemize}
These similarities reflect a common mathematical foundation inherited from cybernetics and descendant fields. We now discuss these two points in more detail.

\subsection{Model Formulation}

The primary distinction in model formulation is the form of the (non-linear) functions parameterizing dependencies. \cite{rao1999predictive} parameterize the conditional likelihood as a linear function followed by an element-wise non-linearity. Friston has considered a wider range of functions, e.g., polynomial \citep{friston2008hierarchical}, however, such functions are rarely learned. VAEs instead parameterize these functions using deep networks with multiple layers. The deep network weights are trained through backpropagation, enabling the wide application of VAEs to various data domains.

Predictive coding and VAEs also consider dependencies within each level. \cite{friston2005theory} uses full-covariance Gaussian densities, with the inverse of the covariance matrix (precision) parameterizing linear dependencies within a level. \cite{rao1999predictive} normalize the observations, modeling linear dependencies within the conditional likelihood. These are linear special cases of the more general technique of normalizing flows \citep{rezende2015variational}: a covariance matrix is an affine normalizing flow with linear dependencies \citep{kingma2016improved}. Normalizing flows have been applied throughout each of the distributions within VAEs (Section \ref{sec: pc normalizing flows}), modeling non-linear dependencies across both spatial and temporal dimensions. These flows are also parameterized by deep networks, providing a flexible, yet general modeling approach.

Related to normalization, there are proposals within predictive coding that the precision of the prior could mediate a form of attention \citep{feldman2010attention}. Increasing the precision of a variable serves as a form of gain modulation, up-weighting the error in the objective function, thereby enforcing more accurate inference estimates. This concept is absent from VAEs. However, as VAEs become more prevalent in interactive settings \citep{ha2018recurrent}, i.e., beyond pure generative modeling, this may become crucial in steering models toward task-relevant perceptual inferences.

Finally, predictive coding and VAEs have both been extended to sequential settings. In predictive coding, sequential dependencies may be parameterized by linear functions \citep{srinivasan1982predictive} or so-called generalized coordinates \citep{friston2008hierarchical}, modeling multiple orders of motion. In extensions of VAEs, sequential dependencies are again parameterized by deep networks, in many cases using recurrent networks \citep{chung2015recurrent, fraccaro2016sequential}. Thus, while the specific implementations vary, in either case, sequential dependencies are ultimately functions, which are subject to design choices.

\subsection{Inference}

Although both predictive coding and VAEs typically use variational inference with Gaussian approximate posteriors, Sections \ref{sec: predictive coding} \& \ref{sec: vaes} illustrate key differences (Figure \ref{fig: pc pc vae comparison}). Predictive coding generally relies on \textit{gradient-based} optimization to perform inference, whereas VAEs employ \textit{amortized} optimization. While these approaches may, at first, appear radically different, hybrid error-encoding inference approaches (Eq.~\ref{eq: error-encoding model}), such as PredNet \citep{lotter2016deep} and iterative amortization \citep{marino2018iterative}, provide a link. Such approaches receive errors as input, as in predictive coding, however, they have learnable parameters, i.e., amortization. In fact, amortization may provide a crucial element for implementing predictive coding in biological neural networks.

Though rarely discussed, hierarchical predictive coding assumes that the inference gradients, supplied by forward connections, can be readily calculated. But, as seen in Section \ref{sec: pc hpc}, the weights of these forward connections are the transposed Jacobian matrix of the backward connections \citep{rao1999predictive}. This is an example of the \textit{weight transport} problem \citep{grossberg1987competitive}, in which the weights of one set of connections (forward) depend on the weights of another set of connections (backward). This is generally regarded as not being biologically-plausible.

Amortization provides a solution to this problem: \textit{learn} to perform inference. Rather than transporting the generative weights to the inference connections, amortization learns a separate set of inference weights, potentially using local learning rules \citep{bengio2014auto, lee2015difference}. Thus, despite criticism from \cite{friston2018does}, amortization may offer a more biologically-plausible inference approach. Further, amortized inference yields accurate estimates with exceedingly few iterations: even a single iteration may yield reasonable estimates \citep{marino2018iterative}. These computational efficiency benefits provide another argument in favor of amortization.

% Error-based inference is an example of the more general concept of negative feedback. As noted earlier, negative feedback was the core concept of cybernetics, which went on to inspire predictive coding. While hierarchical predictive coding has largely focused on perceptual inference in cortex, the principles of negative feedback apply more broadly to neural systems. Indeed, even at the outset of cybernetics, it was clear that cerebellum plays a central role in negative feedback control \citep{wiener1948cybernetics}. From more recent studies of cerebellum and other cerebellum-like structures \citep{ito1998cerebellar, bell2001memory, kennedy2014temporal}, we are beginning to understand how such circuits correct sensorimotor prediction errors. One prominent example is given by the Purkinje cells of the cerebellum, which appear to receive error signals as inputs and output motor corrections. This follows the general paradigm of (iterative) amortization, mapping errors to updates. Casting these neural circuits in terms of amortization, i.e.,~learned negative feedback, may provide insights into how such error-correcting mechanisms can be learned from experience.

Finally, although predictive coding and VAEs typically assume Gaussian approximate posteriors, there is one additional difference in the ways in which these parameters are conventionally calculated. Friston often uses the Laplace approximation\footnote{\normalsize This is not to be confused with a Laplace distribution. The approximate posterior is still Gaussian.} \citep{friston2007variational}, solving directly for the optimal Gaussian variance, whereas VAEs treat this as another output of the inference model \citep{kingma2014stochastic, rezende2014stochastic}. These approaches can be applied in either setting, e.g., \cite{park2019variational}.

\section{Correspondences}
\label{sec: pc correspondences}

Having connected VAEs and predictive coding, we now discuss possible correspondences between machine learning and neuroscience. In Table~\ref{table: biological correspondences}, top-down and bottom-up cortical projections, each mediated by pyramidal neurons, respectively parameterize the generative model and inference updates. Mapping this onto VAEs suggests that \textbf{deep (artificial) neural networks are in correspondence with pyramidal neuron dendrites} (Figure~\ref{fig: pc pyramidal deep network amortization}, Right), or, more specifically, a deep network corresponds to a collection of pyramidal dendrites operating in parallel. Predictive coding also postulates that lateral interneurons parameterize dependencies within variables as inverse covariance matrices. VAEs parameterize these dependencies using more general normalizing flows, suggesting that \textbf{normalizing flows are in correspondence with lateral interneurons}. While normalizing flows also use deep networks, the effect that they have on computation tends to be restricted and simple (e.g.,~affine). We now briefly discuss these correspondences.

% These correspondences are obviously quite coarse-grained, and many details are left to be filled-in. However, they may provide a useful starting point for shifting the current analogies between machine learning and neuroscience. Below, we explore some of the consequences of these correspondences.

% As mentioned above, normalizing flows are a non-linear generalization of linear covariance matrices, suggesting the possibility of non-linear normalization computations in cortex and elsewhere.
% \begin{figure}[t!]
%     \centering
%     \includegraphics[width=0.7\textwidth]{figures/neuron_as_network.pdf}
%     \caption{\textbf{Pyramidal Neurons \& Deep Networks}. Connecting deep latent variable models with predictive coding places deep networks (bottom) in correspondence with the dendrites of pyramidal neurons (top). This is in contrast with conventional one-to-one analogies of biological and artificial neurons, suggesting a larger role for non-linear dendritic computation and alternative correspondences for backpropagation.}
%     \label{fig: pc neuron as network}
% \end{figure}

\subsection{Pyramidal Neurons \& Deep Networks}

\begin{figure}[t!]
    \centering
    \includegraphics[width=0.8\textwidth]{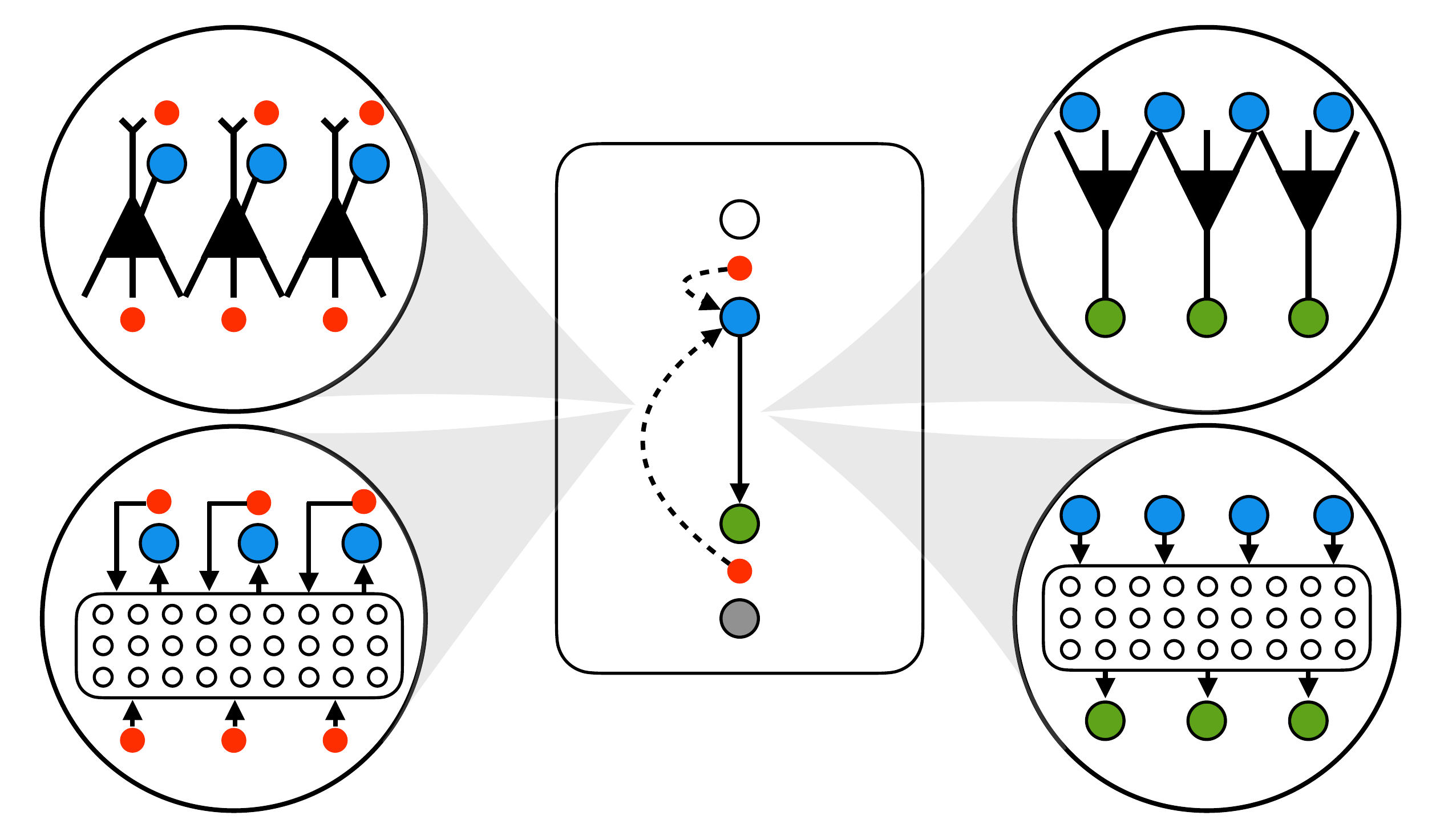}
    \caption{\textbf{Pyramidal Neurons \& Deep Networks}. Connecting VAEs with predictive coding places deep networks (\textit{bottom}) in correspondence with the dendrites of pyramidal neurons (\textit{top}), both for generation (\textit{right}) and (amortized) inference (\textit{left}).}
    \label{fig: pc pyramidal deep network amortization}
\end{figure}

\paragraph{Non-linear Dendritic Computation} Placing deep networks in correspondence with pyramidal dendrites suggests that (some) biological neurons may be better computationally described as \textit{non-linear} functions. Evidence from neuroscience supports this claim. Early simulations showed that individual pyramidal neurons, through dendritic processing, could operate as multi-layer artificial networks \citep{zador1992nonlinear, mel1992clusteron}. This was later supported by empirical findings that pyramidal dendrites act as computational `subunits,' yielding the equivalent of a two-layer artificial network \citep{poirazi2003pyramidal, polsky2004computational}. More recently, \cite{gidon2020dendritic} demonstrated that individual pyramidal neurons can compute the XOR operation, which requires non-linear processing. This is supported by further modeling work \citep{jones2020can, beniaguev2020single}. Positing a more substantial role for dendritic computation \citep{london2005dendritic} moves beyond the simplistic comparison of biological and artificial neurons that currently dominates. Instead, neural computation depends on morphology and circuits.

% Further, rather than assuming every neuron is equivalent, i.e.,~linear summation with non-linearity, separate classes of neurons would represent distinct function classes, likely derived from their morphology. This places a greater emphasis on understanding the particular intricacies of neural circuits, rather than assuming a uniform network of identical computational elements.

% \begin{figure}[t!]
%     \centering
%     \includegraphics[width=0.7\textwidth]{figures/amortization.pdf}
%     \caption{\textbf{Pyramidal Neurons \& Amortization}. In predictive coding, inference updating is implemented using forward pyramidal neurons in cortex, taking prediction errors as input. In deep latent variable models, iterative amortized inference plays a similar role, continuing the analogy of pyramidal neurons and deep networks. Interestingly, this suggests a separation of processing in apical (upper) and basal (lower) dendrites, incorporating errors from the current and lower level.}
%     \label{fig: pc amortization}
% \end{figure}

\paragraph{Amortization} 
Pyramidal neurons mediate both top-down and bottom-up cortical projections. Under predictive coding, this suggests that inference relies on learned, non-linear functions, i.e., \textit{amortization}. One such implementation is through pyramidal neurons with separate apical and basal dendrites, which, respectively, receive top-down and bottom-up inputs \citep{bekkers2011pyramidal, guergiuev2016biologically}. Recent evidence from \cite{gillon2021learning} suggests that these are top-down predictions and bottom-up errors. These neurons may implement iterative amortized inference \citep{marino2018iterative}, separately processing top-down and bottom-up error signals to update inference estimates (Figure~\ref{fig: pc pyramidal deep network amortization}, Left). While some empirical support for amortization exists \citep{yildirim2015efficient, dasgupta2018remembrance}, further investigation is needed. Finally, this perspective implies separate computational processing for prediction and inference, with distinct (but linked) frequencies. While some evidence supports this conjecture \citep{bastos2015visual}, it is unclear how this could be implemented in biological neurons.

\begin{figure}[t!]
    \centering
    \includegraphics[width=0.7\textwidth]{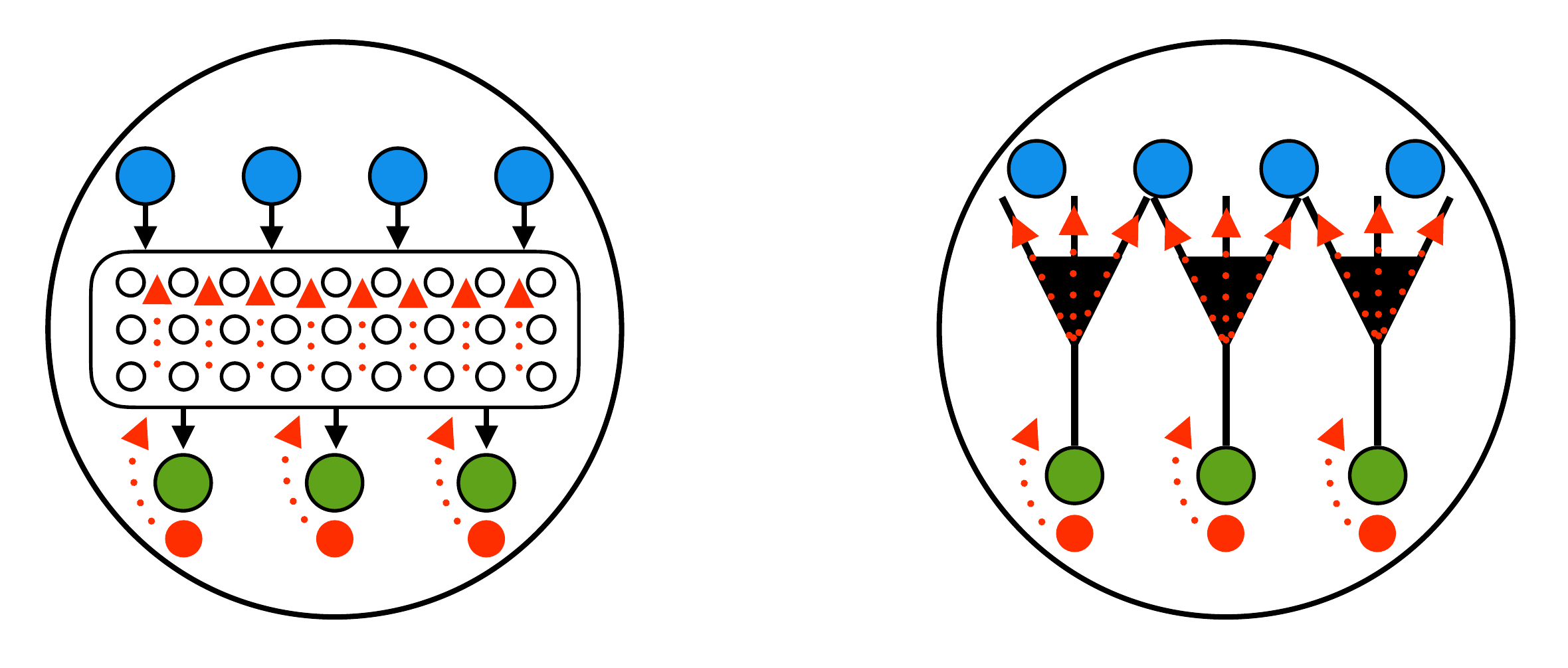}
    \caption{\textbf{Backpropagation Within Neurons}. If deep networks are in correspondence with pyramidal neurons, this implies that backpropagation (\textit{left}) is analogous to learning within neurons, perhaps via backpropagating action potentials (\textit{right}).}
    \label{fig: pc backprop}
\end{figure}

% dendrites suggests an alternative perspective on the biological-plausibility of backpropagation. In deep latent variable models, backpropagation is only performed across variables that are directly connected through a conditional probability (left). From the perspective presented here, this corresponds to learning \textit{within} pyramidal neurons. One possible implementation may be through backpropagating action potentials, perhaps combined with other neuromodulatory inputs (right).

\paragraph{Backpropagation} The biological plausibility of backpropagation is an open question \citep{lillicrap2020backpropagation}. Critics argue that backpropagation requires non-local learning signals \citep{grossberg1987competitive, crick1989recent}, whereas the brain relies largely on local learning rules \citep{hebb1949organization, markram1997regulation, bi1998synaptic}. ``Biologically-plausible'' formulations of backpropagation have been proposed \citep{stork1989backpropagation, kording2001supervised, xie2003equivalence, hinton2007how, lillicrap2016random}, attempting to reconcile this disparity. Yet, consensus is still lacking. From another perspective, the apparent biological implausibility of backpropagation may instead be the result of incorrectly assuming a one-to-one correspondence between biological and artificial neurons.

Placing deep networks in correspondence with pyramidal neurons suggests a different perspective on the biological-plausibility debate. In hierarchical latent variable models, prediction errors at each level provide a local learning signal \citep{friston2005theory, bengio2014auto, lee2015difference, whittington2017approximation}. Thus, learning within each latent level is performed through optimization of local errors. This is exemplified by hierarchical VAEs \citep{sonderby2016ladder}, which utilize backpropagation \textit{within} each latent level, but not \textit{across} levels. This suggests that learning \textit{within} pyramidal neurons may be more analogous to backpropagation (Figure~\ref{fig: pc backprop}). One possible candidate is backpropagating action potentials \citep{stuart1994active, williams2000backpropagation}, which actively propagate a signal of neural activity back to synaptic inputs \citep{stuart1997action, brunner2016analogue}, resulting in a variety of synaptic changes throughout the dendrites \citep{johenning2015ryanodine}. While computational models from \cite{schiess2016somato} support this conjecture, further investigation is needed.
% While many details remain unclear, this overall perspective of backpropagation \textit{within} neurons, rather than across networks of neurons, offers a more biologically-plausible alternative; all signals are local to the pyramidal neurons/dendrites within the cortical circuit. Given the proposed theoretical role and empirical observations, this possible correspondence between backpropagation and backpropagating action potentials warrants further investigation.

\subsection{Lateral Inhibition \& Normalizing Flows}

\paragraph{Sensory Input Normalization} One of the key computational roles of early sensory areas appears to be reducing spatiotemporal redundancies, i.e., normalization. In retina, this is performed through lateral inhibition via horizontal and amacrine cells, removing correlations \citep{graham2006can, pitkow2012decorrelation}. Normalization and prediction are inseparable, and, accordingly, previous works have framed early sensory processing in terms of spatiotemporal predictive coding \citep{srinivasan1982predictive, hosoya2005dynamic, palmer2015predictive}. This is often motivated in terms of increased sensitivity or efficiency \citep{srinivasan1982predictive, atick1990towards} due to redundancy reduction \citep{barlow1961possible, barlow1989finding}, i.e.,~compression. 

If we consider cortex as a hierarchical latent variable model, then early sensory areas are implicated in parameterizing the conditional likelihood. The ubiquity of normalization in these areas is suggestive of normalization in a flow-based model, implementing the inference direction of a flow-based conditional likelihood \citep{agrawal2016deep, winkler2019learning}. In addition to the sensitivity and efficiency benefits cited above, this learned, normalized space simplifies downstream generative modeling and improves generalization \citep{marino2020improving}.

\begin{figure}[t!]
    \centering
    \includegraphics[width=0.8\textwidth]{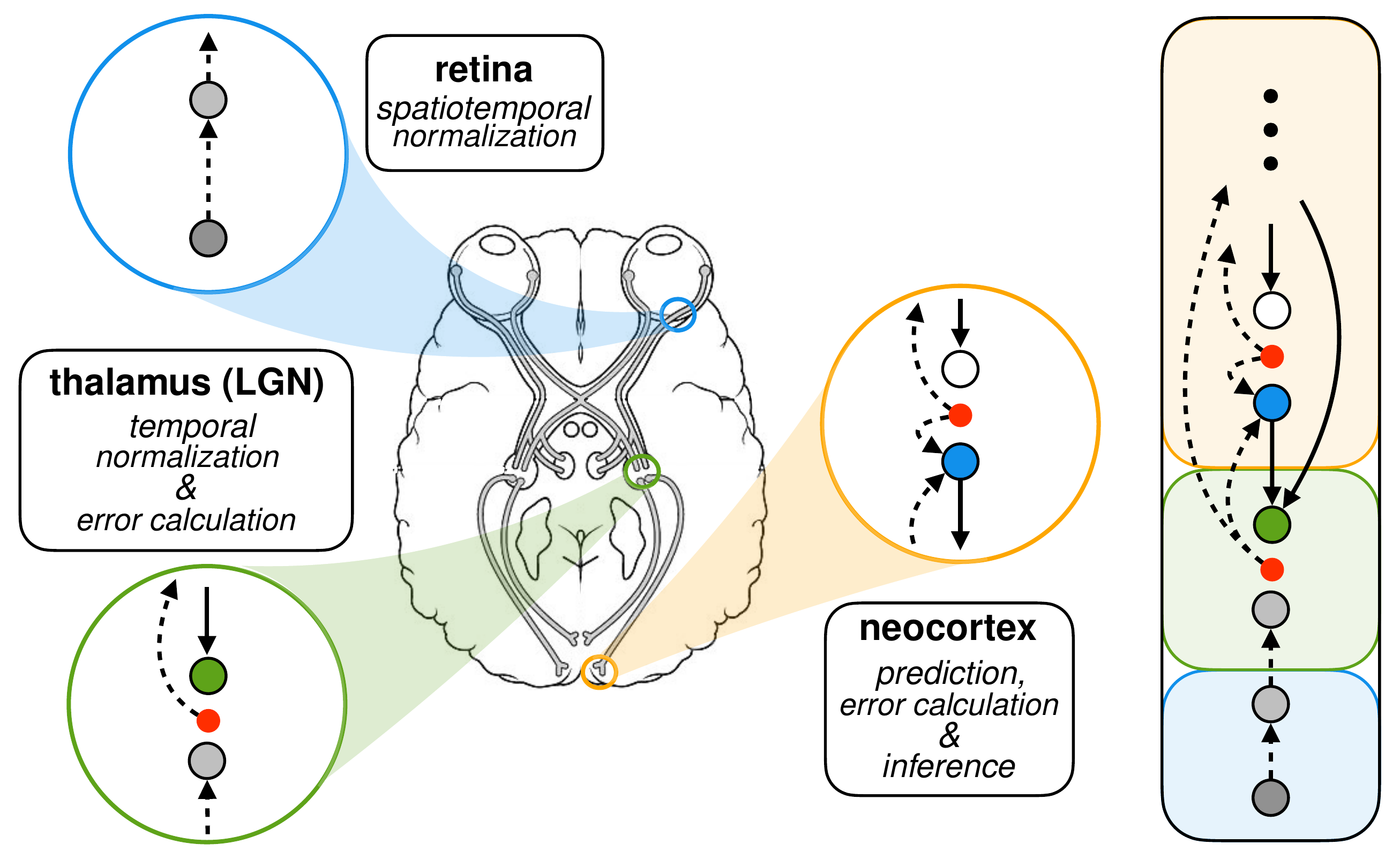}
    \caption{\textbf{Visual Pathway}. Retina and LGN are interpreted as implementing normalizing flows, i.e.,~spatiotemporal predictive coding, reducing spatial and temporal redundancy in the visual input (dashed arrows between gray circles). LGN is also the lowest level for hierarchical predictions from cortex. Using prediction errors throughout the hierarchy, forward cortical connections update latent estimates.}
    \label{fig: pc overall schematic}
\end{figure}

\paragraph{Normalization in Thalamus} 
Normalization also appears to occur in first-order thalamic relays, such as the lateral geniculate nucleus (LGN). \cite{dong1995temporal} framed LGN in terms of temporal normalization, with supporting evidence provided by \cite{dan1996efficient}. This has the effect of removing predictable temporal structure, e.g., static backgrounds. Under the interpretation above, this is an additional inference stage of a flow-based conditional likelihood \citep{marino2020improving}.

% Following the interpretation above, normalization in thalamus may provide a second stage of normalizing flow, further removing redundancy. \cite{marino2020improving} implemented this general technique using autoregressive flows in deep latent variable models. This low-level temporal normalization technique removes static backgrounds, improving modeling performance and generalization. Further work is needed to assess the functional form (affine/non-affine) and types of dependencies (linear/non-linear) implemented in first-order thalamic relays, though suggestions are given by \cite{dong1995temporal}.

\paragraph{Normalization in Cortex}
Normalization, via local lateral inhibition, is also found throughout cortex \citep{king2013inhibitory}.  \cite{friston2005theory} suggested that this plays the role of inverse covariance (precision) matrices, modeling dependencies between dimensions within the same latent level of the hierarchy. This corresponds to parameterizing approximate posteriors \citep{rezende2015variational, kingma2016improved} and/or conditional priors \citep{huang2017learnable} with affine normalizing flows with linear dependencies.\footnote{\normalsize Specifically, \cite{friston2005theory} employs zero-phase component analysis (ZCA) whitening, whereas \cite{kingma2016improved} explored Cholesky whitening.} Normalizing flows offers a more general framework for describing these computations. Further, \cite{friston2005theory} assumes that these dependencies are modeled using \textit{symmetric} weights, whereas normalizing flows permits \textit{non-symmetric} schemes, e.g.,~using autoregressive models \citep{kingma2016improved} or ensembles \citep{uria2014deep}. These weights can also be restricted to local spatial regions \citep{vahdat2020nvae}. Similar normalization operations can also parameterize temporal dynamics \citep{marino2020improving}, akin to Friston's generalized coordinates \citep{friston2008hierarchical}. The overall computational scheme is shown in Figure~\ref{fig: pc overall schematic}.

\section{Discussion}
\label{sec: pc discussion}

We have reviewed predictive coding and VAEs, identifying their shared history and formulations. These connections provide an invaluable link between leading areas of theoretical neuroscience and machine learning, hopefully facilitating the transfer of ideas across fields. We have initiated this process by proposing two novel correspondences suggested by this perspective: 1) dendrites of pyramidal neurons and deep networks, and 2) lateral inhibition and normalizing flows. Placing pyramidal neurons in correspondence with deep networks departs from the traditional one-to-one analogy of biological and artificial neurons, raising questions regarding dendritic computation and backpropagation. Normalizing flows offers a more general framework for normalization via lateral inhibition. Connecting these areas may provide new insights for both machine learning and neuroscience, helping us move beyond overly simplistic comparisons.

\textbf{Predictive Coding $\rightarrow$ VAEs} Although considerable independent progress has recently occurred in VAEs, such models are often still trained on relatively simple, standardized datasets of static images. Thus, predictive coding and neuroscience may still hold insights for improving these models for real-world settings. For instance, the correspondences outlined above may offer new architectural insights in designing deep networks and normalizing flows, e.g., drawing on dendritic morphology, short-term plasticity, and connectivity. Predictive coding has also used prediction precision as a form of attention \citep{feldman2010attention}. More broadly, neuroscience may provide insights into interfacing VAEs with other computations, as well as within embodied agents.

\textbf{VAEs $\rightarrow$ Predictive Coding} Another motivating factor in connecting these areas stems from a desire for large-scale, testable models of predictive coding. While predictive coding offers general considerations for neural activity, e.g., predictions, prediction errors, and extra-classical receptive fields \citep{rao1999predictive}, it is difficult to align such hypotheses with real data due to the many possible design choices \citep{gershman2019does}. Current models are often implemented in simplified settings, with few, if any, learned parameters. VAEs, in contrast, offer a large-scale test-bed for implementing models and evaluating them on natural stimuli. This may offer a more nuanced perspective over current efforts to compare biological and artificial neural activity \citep{yamins2014performance}.

While we have reviewed many topics across neuroscience and machine learning, for brevity, we have focused exclusively on passive perceptual settings. However, separate, growing bodies of work are incorporating predictive coding \citep{adams2013predictions} and VAEs \citep{ha2018recurrent} within active settings, e.g., reinforcement learning. We are hopeful that the connections in this paper will inspire further insight in such areas. 

% TODO: mention identifiability and control briefly

% \paragraph{Control} For practical reasons, we have almost exclusively focused on perception in discussing predictive coding and the ideas in this paper. This is a result of the fact that predictive coding was initially developed and studied in the context of generative models of sensory inputs \citep{srinivasan1982predictive, rao1999predictive, friston2005theory}. However, if the cortical microcircuit implements a general-purpose modeling and inference algorithm, we should expect similar computations to be applicable to motor and prefrontal cortices. Over the past decade, Friston and colleagues have developed a range of ideas in this direction, interpreting motor control as a process of proprioceptive prediction \citep{adams2013predictions} and prefrontal cortex as performing hierarchical goal inference \citep{pezzulo2018hierarchical}. Such ideas can be seen as modern extensions of early ideas in cybernetics \citep{wiener1948cybernetics, mackay1956epistemological, powers1973behavior}, using motor control to correct errors between desired and actual outcomes. Indeed, Friston's efforts are part of a resurgence of interest in framing control as probabilistic inference \citep{levine2018reinforcement}, in which the same inference algorithms can be applied to policy optimization \citep{marino2020iterative}. Nevertheless, despite recent progress and perspectives in this area \citep{hafner2020action}, further work is needed to combine perception and control under a unified formulation.

\subsection*{Acknowledgments}
Sam Gershman and Rajesh Rao provided helpful comments on this manuscript, and Karl Friston engaged in useful early discussions related to these ideas. We also thank the anonymous reviewers for their feedback and suggestions.

\bibliographystyle{apacite}
\bibliography{references}
\newpage
\appendix
\section{Variational Bound Derivation}
\label{app: elbo deriv}
We can express the KL divergence between $q (\mathbf{z} | \mathbf{x})$ and $p_\theta (\mathbf{z} | \mathbf{x})$ as
\begin{align}
    D_\textrm{KL} (q (\mathbf{z} | \mathbf{x}) || p_\theta (\mathbf{z} | \mathbf{x})) & = \mathbb{E}_{\mathbf{z} \sim q (\mathbf{z} | \mathbf{x})} \left[ \log q (\mathbf{z} | \mathbf{x}) - \log p_\theta (\mathbf{z} | \mathbf{x}) \right]  \nonumber \\
    & = \mathbb{E}_{\mathbf{z} \sim q (\mathbf{z} | \mathbf{x})} \left[ \log q (\mathbf{z} | \mathbf{x}) - \log \left( \frac{p_\theta (\mathbf{x} , \mathbf{z})}{p_\theta (\mathbf{x})} \right) \right]  \nonumber \\
    & = \underbrace{\mathbb{E}_{\mathbf{z} \sim q (\mathbf{z} | \mathbf{x})} \left[ \log q (\mathbf{z} | \mathbf{x}) - \log p_\theta (\mathbf{x} , \mathbf{z}) \right]}_{- \mathcal{L} (\mathbf{x}; q, \theta)} + \log p_\theta (\mathbf{x}).
\end{align}

% \begin{thebibliography}{100}
% \providecommand{\natexlab}[1]{#1}
% \expandafter\ifx\csname urlstyle\endcsname\relax
%   \providecommand{\doi}[1]{doi:\discretionary{}{}{}#1}\else
%   \providecommand{\doi}{doi:\discretionary{}{}{}\begingroup
%   \urlstyle{rm}\Url}\fi

% \end{thebibliography}
\end{document}